\documentclass[10pt,twocolumn,letterpaper]{article}
\usepackage[pagebackref=true,breaklinks=true,letterpaper=true,colorlinks,bookmarks=false]{hyperref}
\usepackage{cvpr}
\usepackage{times}
\usepackage{epsfig}
\usepackage{graphicx}
\usepackage{amsmath}
\usepackage{amssymb}
\usepackage{algorithmic}
\DeclareRobustCommand\onedot{\futurelet\@let@token\@onedot}
\def\etal{\emph{et al}.}

\cvprfinalcopy
\def\cvprPaperID{2783} 

\ifcvprfinal\fi
\begin{document}

\title{Deep, Dense, and Low-Rank Gaussian Conditional Random Fields}
\author{Siddhartha Chandra \hfill Iasonas Kokkinos\\
{\tt \small siddhartha.chandra@inria.fr} \hfill {\tt \small  i.kokkinos@cs.ucl.ac.uk}\\
{\small {INRIA GALEN} \& Centrale Sup\'elec Paris, France \hspace{20mm} University College London, U.K.}\\
}
\maketitle

\begin{abstract}
In this work we introduce a fully-connected graph structure in the Deep Gaussian Conditional Random Field (G-CRF) model. For this 
we express the pairwise interactions between pixels as the inner-products of low-dimensional embeddings, delivered by a new subnetwork of a deep architecture. We efficiently minimize the resulting energy by solving the resulting low-rank linear system with conjugate gradients, and derive an analytic expression for the gradient of our embeddings which allows us to train them end-to-end with backpropagation. 

We demonstrate the merit of our approach by achieving state of the art results on three challenging Computer Vision benchmarks, namely semantic segmentation, human parts segmentation, and saliency estimation. Our implementation is fully GPU based, built on top of the Caffe library, and will be made publicly available.
\end{abstract}

\section{Introduction}
\label{sec:Introduction}

\begin{figure}[t]
 \includegraphics[width=\linewidth]{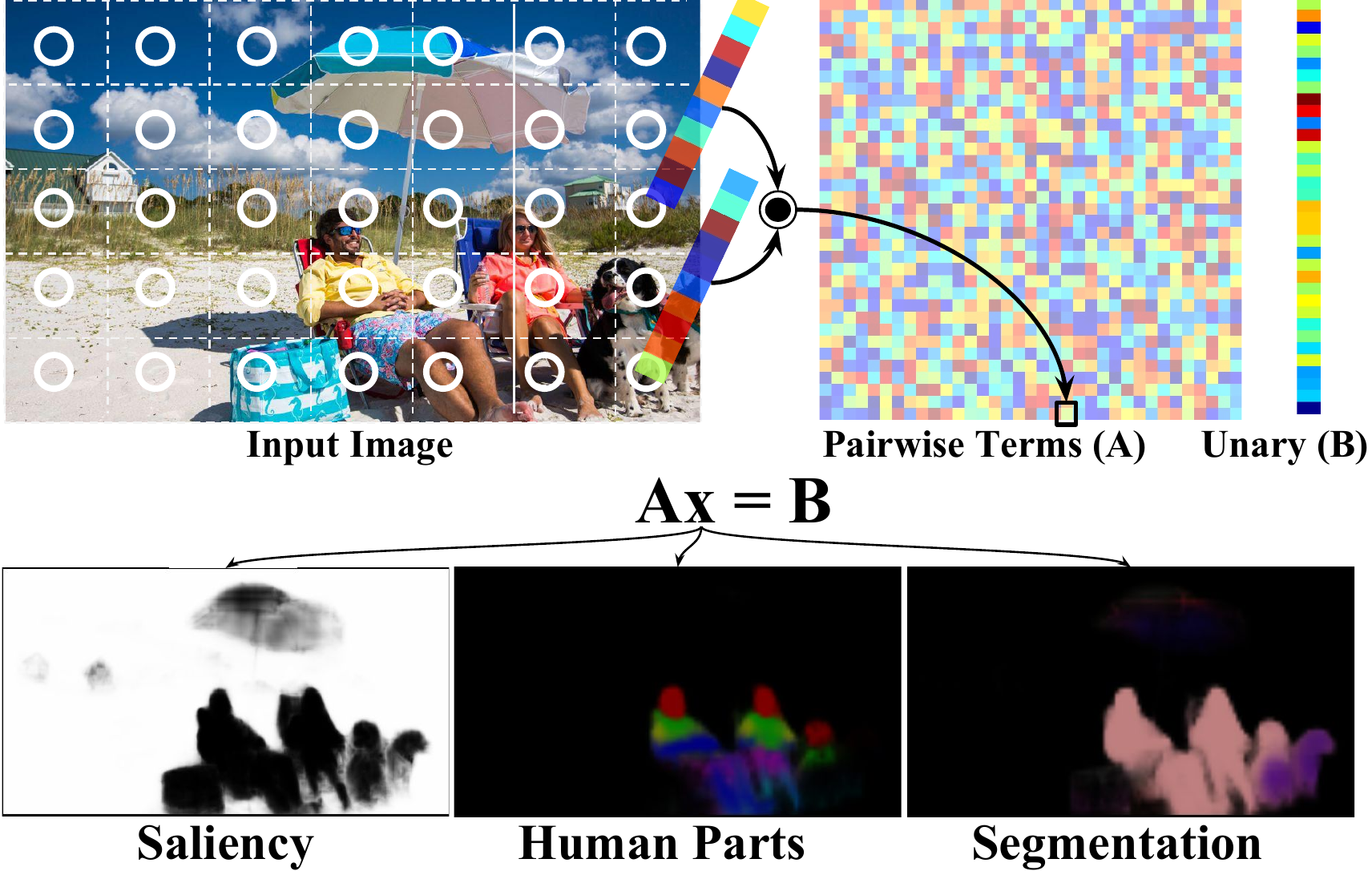}
 \caption{A brief overview of our proposed fully-connected model for dense labeling tasks. 
Each patch in the input image denotes a node in our fully-connected graph structure.
While the unary terms (B) are obtained from 
the output of a fully convolutional network,
our framework expresses the pairwise terms (A) as dot products of pixel embeddings. These embeddings are also obtained from the output of a parallel stream
of the fully convolutional network. We solve a system of linear equations $A\textbf{x} = B$, to infer the prediction $\textbf{x}$. Our approach can
be used to learn task-specific unary and pairwise terms for dense labeling tasks such as semantic segmentation, part 
segmentation and saliency estimation.} 
 \label{fig:teaser}
\end{figure}

Structure prediction combined with deep-learning has delivered striking results on a variety of computer vision benchmarks~\cite{gcrf,deeplab1,deeplab2,schwing,Vemulapalli_2016_CVPR,vu2015context,crfrnn}.
Rather than leaving all of the modeling task to a generic deep network, the explicit treatment of interactions between image regions in a principled manner has been advocated in ~\cite{nonparam,schwing,densecrf,vv,crfrnn}.
Several of the earlier approaches~\cite{deeplab1,deeplab2} exploited the Dense-CRF~\cite{densecrf} framework that uses a pre-determined parametric pairwise interaction term among output variables, effectively  refining  object boundaries while compensating for the effects of spatial downsampling within the network.
Other more recent approaches \cite{Adelaide,depthcrf,DPN,Vemulapalli_2016_CVPR,vu2015context,crfrnn} showed that structure prediction could be done in an end-to-end
manner for both sparsely-connected~\cite{Vemulapalli_2016_CVPR} and fully-connected \cite{Adelaide,DPN,vu2015context,crfrnn} graphical structures.
These approaches typically took a mean-field approximation to the original CRF as in ~\cite{densecrf}
and implemented mean-field inference for a fixed number of iterations via back-propagation-through-time.
Even though some of these approaches allowed for end-to-end training for fully-connected graphical models,
they all relied on approximate inference.

\begin{figure*}[t!]
\centering
 \includegraphics[width=0.95\textwidth]{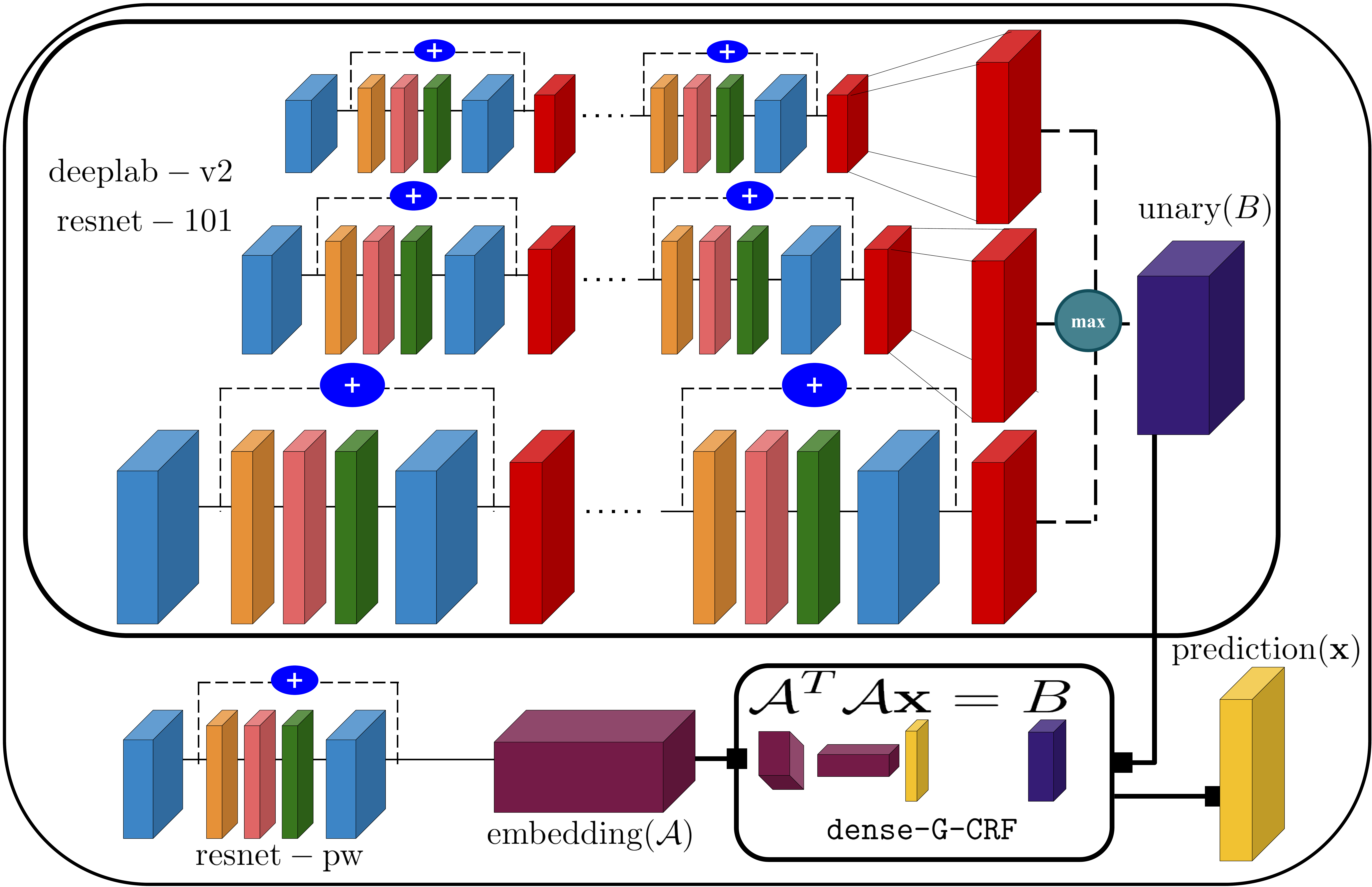}
 \caption{A detailed schematic representation of our fully convolutional neural network with the  \texttt{dense-G-CRF} module.
 Our base network for generating the unary terms is deeplab-v2-resnet-101. This network has three parallel branches of the
 resnet-101 network operating at different image scales. The final unary scores are obtained by upsampling the network responses
 for the low resolution branches to the original resolution, and taking an element-wise maximum of the responses at the three resolutions. Our pairwise terms are generated by
 a parallel pairwise residual network branch (resnet-pw) which has layers \texttt{conv-1} to \texttt{res4a} of the resnet-101 network. The resnet-pw network
 outputs the pixel embeddings of our formulation. The unary terms and pairwise embeddings are then fed to our fully connected G-CRF
 module (\texttt{dense-G-CRF}). This module outputs the prediction $\textbf{x}$ by solving the inference equation $\mathcal{A}^T \mathcal{A} \textbf{x} = B$.
 Additionally, the \texttt{dense-G-CRF} module computes the error gradients for the unary and pairwise terms and
 and back propagates them through the respective streams of the network during the training phase.
 }
 \label{fig:full}
\end{figure*}

More recently, drawing inspiration from \cite{rtf,TappenLAF07} the deep G-CRF model~\cite{gcrf} alleviated the need for approximate inference.
In particular, given an image $I$ the G-CRF model defines a joint posterior distribution through a Gaussian multi-variate density:
\begin{equation*}
p(\textbf{x}|I) \propto \exp(-\frac{1}{2}\textbf{x}^T A_{I} \textbf{x} + B_{I} \textbf{x})\text{,}
\end{equation*}
where $A_I$,$B_I$ are the canonical parameters of the data-dependent Gaussian density. Dropping the dependence on $I$ for simplicity, we note that 
 $A$ can be seen as the \emph{inverse covariance} or \emph{precision} matrix, and for a positive-definite matrix $A$, inference involves solving the system of linear equations $A \textbf{x} = B$.
The deep G-CRF model showed that the parameters of the Gaussian density could be learnt in an end-to-end manner through a CNN, and exact inference could be made efficiently
by using the conjugate gradients~\cite{conjugategradient} algorithm.
While the G-CRF model allowed for efficient and exact inference,
in practice it can only 
capture interactions in small ($4-$,$8-$ and $12-$connected) neighbourhoods, which result in a sparse
precision matrix. The authors take advantage of this by using sparse linear algebra optimizations to provide
a fast implementation. While this speed is preferable, the model loses some of its richness by ignoring 
long range interactions.

To recapitulate, on one hand we have approaches that employ approximate inference for potentially fully-connected graphical structures, and on the other we have an approach that allows exact inference but only for sparsely-connected graphical structures. 

In this work, we propose a strategy that enables efficient
and exact inference for fully-connected graphical models. We do so by 
extending the G-CRF model to fully connected graphical models.

The extension to fully-connected graphs is technically challenging because of the non-sparse precision matrix it involves.
Assuming an image size of $500\times 500$ pixels, $21$ labels (\texttt{PASCAL VOC} benchmark),
and a network with a spatial downsampling factor of $8$ (as in \cite{deeplab1,deeplabv2}), the number of pairwise terms would be $\left(\frac{500}{8} \times \frac{500}{8} \times 21 \right)^2 \sim 10^{10}$.
This is prohibitively large due to both memory and computational requirements.
To overcome this challenge, we advocate using a \emph{low-rank} precision matrix. In particular we propose
\emph{composing} the precision matrix $A$ via dot-products of ``low-dimensional'' pixel embeddings $\mathcal{A}$, that is $A = \mathcal{A}^T \mathcal{A}$.

This drastically reduces the memory footprint. Rather than explicitly obtaining the precision matrix, as in the original G-CRF work, 
we now only need to discover and store in memory low-dimensional pixel embeddings. While maintaining a fully-connected graph structure, we can tune the embedding dimension  at will, so as to meet the available resources and/or accuracy standards. 
Furthermore, the embedding's dimensionality can be used to control the computational complexity of solving the resulting linear system with Conjugate Gradients, since the latter   operate with $\mathcal{A}$
rather than $A$.

Figure \ref{fig:teaser} shows the overview of our approach. The input image first passes through a fully convolutional network,
which results in spatial downsampling by a factor of $8$.
Each $8\times8$ patch in the input image is represented by a node that is accompanied by the dotted lines.
The network generates the unary terms and the pixel embeddings. The pairwise terms between
any two nodes are expressed as the dot-product of their respective embeddings, yielding the precision matrix $A$. Given the unary terms $B$ and the pairwise terms $A$,
we infer the prediction $\textbf{x}$ by solving $A\textbf{x} = B$.
Solving this system of linear equations results in coupling among all the node variables.
 This approach is \emph{loss-agnostic} and works with arbitrary
differentiable losses, and has enabled us to learn task-specific networks
for a variety of tasks, such as saliency estimation, human parts segmentation, and semantic segmentation.
Finally, our implementation is efficient, fully GPU based, and built on top of the \emph{Caffe} library~\cite{jia2014caffe}.

We first give a brief review of the G-CRF model in Sec. \ref{sec:gcrf}, then provide a 
 detailed description of our approach in Sec. \ref{sec:kernel}, and finally  demonstrate the merit of our approach on three challenging tasks, namely, semantic segmentation (Sec. \ref{sec:seg}),
human parts segmentation (Sec. \ref{sec:parts}), and saliency estimation (Sec. \ref{sec:sal}).

\section{Deep Gaussian Conditional Random Fields}
\label{sec:gcrf}
We briefly describe the Deep G-CRF formulation of \cite{gcrf} for completeness,  following  the notation of \cite{gcrf}. 

We consider an image containing $P$ pixels where
each pixel can take a label $l \in \{1,\ldots,L\}$. The predictions are represented as a real-valued vector, 
 $\textbf{x} \in \mathbb{R}^{PL}$, giving the score for every pixel-label combination. 
The unary and pairwise terms are respectively denoted by $B \in \mathbb{R}^{PL}$, $A \in \mathbb{R}^{PL \times PL}$.
Dropping the probabilistic formulation provided in the introduction, we  view the G-CRF  as a structured layer \cite{Sminchisescu} that minimizes a quadratic energy  $E({\bf x})$:
\begin{equation}
 E({\bf x}) = \frac{1}{2} {\bf x}^T (A+ \lambda \textbf{I}) {\bf x} - B{\bf x},
 \label{eqn:energy}
\end{equation}
where a positive constant $\lambda$ is added to the diagonal entries of $A$ to make it positive definite.
Given $\lambda$ such that $A + \lambda \mathbf{I}\succ 0$ and any $B$, $E({\bf x})$ has a unique global minimum at 

\begin{equation}
 ( A + \lambda \mathbf{I} ) \mathbf{x} = B\text{.}
 \label{eqn:linearSolver}.
\end{equation}
Thus the ``exact'' inference for $\textbf{x}$ that gives the minimum energy involves solving a system of linear equations.

The authors in \cite{gcrf} propose learning the model parameters $A$, and $B$ via end-to-end network training. 
To achieve this, they design a G-CRF \emph{layer} or deep-learning module, which (a) collects the unary and pairwise terms
during the forward pass and outputs the predictions, and (b) collects the gradients of the prediction in the backward pass
and uses them to compute the gradients of the model parameters, which it then back propagates through the network.

In particular, considering that the G-CRF \emph{layer}  obtains a gradient  for the loss $\mathcal{L}$ with respect to its output $\textbf{x}$,
$\frac{\partial \mathcal{L}}{\partial \textbf{x}}$,
the authors of \cite{gcrf}  show that
the gradients of
the unary terms $ \frac{\partial \mathcal{L}}{\partial B}$ can be obtained by solving a new system of linear equations:

\begin{equation}
 ( A + \lambda \mathbf{I} ) \frac{\partial \mathcal{L}}{\partial B} = \frac{\partial \mathcal{L}}{\partial \textbf{x}}\text{,}
 \label{eqn:dldbres}
\end{equation}
%
while  the gradients of the pairwise terms $ \frac{\partial \mathcal{L}}{\partial A}$ are given by:
\begin{equation}
 \frac{\partial \mathcal{L}}{\partial A} = - \frac{\partial \mathcal{L}}{\partial B}  \otimes \mathbf{x}\text{,}
 \label{eqn:dlda2_1}
\end{equation}
where $\otimes$ denotes the Kronecker product operator.

\section{Low-Dimensional G-CRF Pixel Embeddings}
\label{sec:kernel}
Having introduced more formally the G-CRF model,
we now  establish the fully-connected G-CRF through low-dimensional embeddings. We use the same mathematical notation as in the previous section.

For brevity, we denote the number of variables in our formulation by $N$, where $N = P\times L$. We also denote by $D$ the dimensionality of the
feature embedding space, where $D << N$. We denote by 
$\mathcal{A} \in \mathbb{R}^{D \times N}$ the matrix of pixel embeddings for all variables in our formulation.
We begin by first defining the pairwise terms or the precision matrix $A$ in terms of the pixel embeddings as:
\begin{equation}
 A = \mathcal{A}^T \mathcal{A} \text{.}
 \label{eqn:kerneldef}
\end{equation}

The matrix $A \in \mathbb{R}^{N\times N}$ gives the pairwise terms for every pair of pixels and labels in the label set, i.e. 
\begin{eqnarray*}
A(p_i,p_j,l_m,l_n) = \mathcal{A}(p_i,l_m)^T\mathcal{A}(p_j,l_n),\\ i,j \in \{ 1,\ldots,P \}, \hspace{5mm} m,n \in \{ 1,\ldots,L \}.
\end{eqnarray*}
Since $A$ is composed via dot-products of pixel embeddings,
it is symmetric and positive semi definite by design. 
However, $A$ is low-rank with a rank of $D$,
and not strictly positive definite. 
Therefore, to make it strictly positive definite, we add a positive constant $\lambda$ to its diagonal
elements. We note that $A = \mathcal{A}^T \mathcal{A}+ \lambda \textbf{I}$ is now guaranteed to be positive definite for any $\lambda$, unlike the case for the Sparse G-CRF, where $\lambda$ had to be set by hand in \cite{gcrf}. 

Following this definition of $A$, the energy function that we wish to minimize now is given by:
\begin{equation}
 E({\bf x}) = \frac{1}{2} {\bf x}^T (\mathcal{A}^T \mathcal{A}+ \lambda \textbf{I}) {\bf x} - B{\bf x}\text{.}
 \label{eqn:kernelenergy}
\end{equation}

Given that the precision matrix is strictly positive definite, $E(\textbf{x})$ has a unique global minimum at
\begin{equation}
 ( \mathcal{A}^T \mathcal{A} + \lambda \mathbf{I} ) \mathbf{x} = B\text{.}
 \label{eqn:kernelinference}
\end{equation}

Thus, inference involves solving a system of linear equations. We take advantage of the positive definiteness of $A$
to use fast Conjugate Gradients method~\cite{conjugategradient} for solving the system of linear equations iteratively. The computational complexity of this method can be controlled by $D$, since we only need to employ vector-matrix products between $\mathbf{x}$ and $\mathcal{A}$, rather than $A$ - resulting in very efficient solvers. 

\subsection*{Gradients of the dense G-CRF parameters}
We now turn to learning the model parameters via end-to-end network training.
To achieve this we require derivatives of the overall loss $\mathcal{L}$ with respect to the model parameters,
namely $\frac{\partial \mathcal{L}}{\partial \mathcal{A}}$ and $\frac{\partial \mathcal{L}}{\partial B}$. 
As described in Eq.~\ref{eqn:kernelinference}, we have an analytical closed form relationship between our model parameters $\mathcal{A}$,$B$, and
the prediction $\textbf{x}$. Therefore, by applying the chain rule of differentiation, we can analytically express the gradients of the model
parameters in terms of the gradients of the prediction. The gradients of the prediction are delivered by the neural network layer on top of 
our \texttt{dense-G-CRF} module through  back- propagation.

The gradients
of the unary terms are straightforward to obtain by substituting Eq.~\ref{eqn:kerneldef} in Eq.~\ref{eqn:dldbres} as:

\begin{equation}
 ( \mathcal{A}^T\mathcal{A} + \lambda \mathbf{I} ) \frac{\partial \mathcal{L}}{\partial B} = \frac{\partial \mathcal{L}}{\partial \textbf{x}}\text{.}
 \label{eqn:kerneldldbres}
\end{equation}
We thus obtain the gradients of the unary terms by solving a system of linear equations.

Turning to the gradients of the pixel embeddings,  $\mathcal{A}$, we use the chain rule of differentiation as follows:
\begin{equation}
 \frac{\partial \mathcal{L}}{\partial \mathcal{A}} =  \left(\frac{\partial \mathcal{L}}{\partial A}\right)  \left(\frac{\partial A}{\partial \mathcal{A}}\right) = \left(\frac{\partial \mathcal{L}}{\partial A}\right)  \left(\frac{\partial }{\partial \mathcal{A}}\mathcal{A}^T \mathcal{A}
 \right)\text{.}\label{eqn:kernelextend}
\end{equation}
We know the expression for $\frac{\partial \mathcal{L}}{\partial A}$ from Eq.~\ref{eqn:dlda2_1}, but to obtain the expression for $\frac{\partial }{\partial \mathcal{A}}\mathcal{A}^T \mathcal{A}$, we need to follow some more tedious steps, which however lead to a simple solution. 

As in \cite{matcalc}, we define a permutation matrix $T_{m,n}$ of size $mn \times mn$ matrix   as follows:
\begin{equation}
 T_{m,n} \text{vec}(M) = \text{vec}(M^T)\text{,}
\end{equation}
where vec$(M)$ is the vectorization operator that vectorizes a matrix $M$ by stacking its columns. 
When premultiplied with another matrix, $T_{m,n}$  rearranges the ordering of rows of that matrix, while when postmultiplied with another matrix, $T_{m,n}$ rearranges its columns.
Using this matrix, we can form the following expression \cite{matcalc}:
\begin{equation}
 \frac{\partial }{\partial \mathcal{A}}\mathcal{A}^T \mathcal{A} = \left({\bf I} \otimes \mathcal{A}^T\right) + \left(\mathcal{A}^T \otimes {\bf I}\right) T_{D,N} \text{,}
 \label{eqn:aat}
\end{equation}
where ${\bf I}$ is the $N\times N$ identity matrix. Substituting Eq.~\ref{eqn:dlda2_1} and Eq.~\ref{eqn:aat} into Eq.~\ref{eqn:kernelextend}, we obtain
\begin{equation}
 \frac{\partial \mathcal{L}}{\partial \mathcal{A}} = - \left(\frac{\partial \mathcal{L}}{\partial B}  \otimes \mathbf{x}\right)\left( \left({\bf I} \otimes \mathcal{A}^T\right) + \left(\mathcal{A}^T \otimes {\bf I}\right) T_{D,N}\right)\label{eqn:kerneldlda}
 \end{equation}

Thus, the gradients of $\mathcal{A}$ can be obtained analytically from the closed form expression in Eq.~\ref{eqn:kerneldlda}.
Despite the apparently complex form, this   final expression is particularly simple
to implement.

We now give an intuitive interpretation of this expression.
As shown in Eq.~\ref{eqn:kernelextend}, the gradients of $\mathcal{A}$ come from the product of two terms.
The first term involves $\frac{\partial \mathcal{L}}{\partial A}$. The quantities $A$ and $B$ can be understood to
be linearly related via $\textbf{x}$, as seen in Eq.~\ref{eqn:linearSolver}. Thus, 
the gradients of $A$ depend linearly on the gradients of $B$, scaled by the coefficients in $\textbf{x}$.
The second term involves $\frac{\partial }{\partial \mathcal{A}}\left(\mathcal{A}^T \mathcal{A}\right)$. This involves
taking the
derivative of a quadratic form of $\mathcal{A}$, i.e. $\mathcal{A}^T\mathcal{A}$ with respect to $\mathcal{A}$. Thus
the second term is linear in $\mathcal{A}$, albeit with repetitions and permutations to satisfy the rules of matrix multiplications.
We use this insight to very efficiently implement the computation of the gradients during the backward pass, 
without computing any Kronecker products or multiplications with $T_{m,n}$ operators explicitly.

%
%
\subsection*{Implementation and Efficiency}
Our approach is implemented as a layer in the Caffe deep learning library~\cite{jia2014caffe},
and we use the conjugate gradients algorithm for solving the systems of linear equations. The complexity
of our method is directly proportional to the size of the matrix of pixel embeddings, $\mathcal{A} \in \mathbb{R}^{D\times PL}$,
thus depends on the embedding dimension, the number of pixels, and the number of labels.
Our implementation is efficient, and exploits the fast linear algebra routines of CUDA blas library.
For these timing comparisons, we use a \texttt{GTX-1080} GPU.
Our inference procedure takes $0.029$s on average 
for the semantic segmentation task ($21$ labels) for an image patch of size $321 \times 321$ pixels
downsampled by a factor of $8$, and for an embedding dimension of $128$. 
This is an order of magnitude faster than the approximate dense CRF mean-field inference which takes
$0.2$s on average. 
The sparse G-CRF, and the Potts-type 
sparse G-CRF from~\cite{gcrf} take $0.021$s and $0.003$s respectively for the same input size.
Thus, our dense inference procedure comes at negligible extra cost compared to the sparse G-CRF.
The average inference times for the same sized input, and $128-$dimensional embeddings for the human parts
segmentation ($7$ labels), and for the saliency estimation task ($2$ labels) are $0.015$s and $0.009$s respectively.
We intend to make our implementation publicly available.

\section{Experiments and Results}
In this section, we describe our experimental setup and results. 

\textbf{Base network.}
Our base network is deeplab-v2 resnet-101~\cite{deeplabv2}, which is a three branch multiresolution network
based on the resnet-101 network~\cite{resnet}. It processes the input image at $3$ resolutions, with scaling factors of $1, 0.75, 0.5$
and then combines the network responses by upsampling them to the original image resolution and taking an element-wise maximum of the responses at the three resolutions.
It also uses atrous spatial pyramid pooling (ASPP), i.e. multiple parallel filters with different dilation 
parameters to better exploit multi-scale features. While this gives a slight performance boost for semantic
segmentation, it does not help performance on other tasks, namely human part segmentation and saliency estimation.
Therefore, we do not use ASPP for these two tasks. The training regime uses random horizontal flipping, and
random scaling of the input image to achieve data augmentation.

\textbf{Fully-Connected G-CRF network.}
Our fully-connected G-CRF (\texttt{dense-G-CRF}) network is shown in figure \ref{fig:full}. The \texttt{dense-G-CRF} network uses the base network
to provide unaries, and a piece of the resnet-101 network in parallel to the base network to construct the pixel embeddings for
the pairwise terms. Our validation experiments on the image segmentation benchmark 
in Sec.\ref{sec:seg} indicate that chopping the pairwise branch, referred to as resnet-pw at
the \texttt{res4a} layer yields the best performance, and we use the same resnet-pw for the other benchmarks. The resnet-pw processes the image
at the original resolution. We use a $2-$phase piece-wise training strategy. We first train the unary network without the pairwise stream, and then 
train the pairwise stream of the network in a second phase, while keeping the unary stream fixed. 
This strategy gives us a smaller training loss than training both unary and pairwise streams at the same time. Each training phase uses $20$K iterations
with a batch size of $10$ as in \cite{deeplabv2}. The initial learning rate for the unary stream is fixed to $0.001$, while for the second phase we set it
to $2.5e-4$. We use a polynomial decaying learning rate with power$=0.9$ as in \cite{deeplabv2}. Training each network takes around $2$ days on a \texttt{GTX-1080} GPU with $8$GB RAM.

\subsection{Semantic Segmentation}
\label{sec:seg}
 Semantic image segmentation task
involves classifying each pixel in the image as belonging to one of a set of candidate classes. In this section, we use our dense labeling framework on the challenging task of semantic image segmentation. We begin with the description of the dataset, followed by a 
brief description of our baselines, before discussing our results.

\textbf{Dataset.} We use our approach on the \texttt{PASCAL VOC 2012} image segmentation benchmark.
This dataset contains
$1464$ training and $1449$ validation images with manually annotated per-pixel labels for twenty foreground object classes, and one background class.
We also use the additional per-pixel segmentation annotations
provided by \cite{hariharan}, obtaining $10582$ training images in total. This benchmark has a \emph{secret} test set containing $1456$ unannotated images.
While the test set images are publicly available, the ground-truth is not available, and the evaluation happens on an online server.
The evaluation
criterion is the pixel intersection-over-union (IOU) metric, averaged across the $21$ classes.

\textbf{Ablation Studies.} We first delve into the ablation studies on the validation set. For these experiments, we use the $10582$ training images, and evaluate on 
the $1449$ validation images. We study the effect of varying the depth of the pairwise network stream by chopping the resnet-101 at three lengths, indicated by the 
standard resnet layer names. Additionally, we also study the effect of varying the size of the pixel-embeddings for the pairwise terms. These results are reported in table \ref{table:seg1}. We notice that the best results are obtained at the embedding dimension of $128$, and the results improve as we increase the depth
of resnet-pw. We do not increase the depth of the resnet-pw beyond \texttt{res4a} due to memory constraints. The improvement over the base-network is $0.75\%$ in mean IoU. 
\begin{table}[!h]
\centering
\begin{tabular}{l |c|c|c|c}
\hline
Base network~\cite{deeplabv2}  &\multicolumn{4}{c}{$76.30$} \\ \hline
\hline
\texttt{dense-G-CRF}&\multicolumn{4}{c}{{Embedding Dimension $\rightarrow$}}\\ \cline{1-1}
resnet-pw size $\downarrow$& \texttt{64} & \texttt{128} & \texttt{256} & \texttt{512} \\ \hline
\texttt{res2a} & $76.79$& $76.81$ & $76.80$ & $76.80$\\ \hline
\texttt{res3a} & $76.98$& $76.85$ & $76.84$ & $76.71$\\ \hline
\texttt{res4a} & $76.95$& $\textbf{77.05}$ & $76.95$ & $76.97$\\ \hline
\end{tabular}
\vspace{2mm}
\caption{Ablation study- mean Intersection Over Union (IOU) accuracy on PASCAL VOC 2012 validation set. We compare the performance of our method against that of the base network, and study the effect of varying the depth of the pairwise stream network, and the size of pixel embeddings.\label{table:seg1}}
\end{table}

\textbf{Performance on test set.}
We now compare our approach with the base network~\cite{deeplabv2},
the base network with the sparse deep G-CRF from \cite{gcrf}, as well as other leading approaches on this benchmark. In these experiments, we train with the augmented training and validation sets, and evaluate performance on the test set. We use our best configuration (\texttt{res4a}, $128$) from table \ref{table:seg1}.

\textbf{Baselines.} The mainstream approach on this task is to use fully convolutional networks~\cite{deeplab1,deeplabv2,fcnn} trained with the 
Softmax cross-entropy loss. For this task, we compare our approach with the state of the art methods on this benchmark. 
The baselines include (a) the multi-scale deeplab network~\cite{deeplab1} which introduced the hole algorithm in conjunction with
dense CRF postprocessing~\cite{densecrf} and used weak supervision to exploit additional training samples with bounding box annotations, (b) the CRF as RNN network~\cite{crfrnn} which proposed first that dense CRF parameters
could be learnt alongside the unary features in an end-to-end manner, (c) the Deeplab+Boundary network~\cite{edg} which exploits an
edge detection detection network to boost the performance of the deeplab network, 
(d) the Adelaide Context network~\cite{Adelaide}
and (e) the deep parsing network both of which learn pairwise terms using variants of the mean-field inference (f) our base network, i.e.
the deeplab-v2 network~\cite{deeplabv2} and (h) the G-CRF model~\cite{gcrf} which combines the deeplab-v2 network with 4-connected Potts type
pairwise terms.

\begin{figure}
\begin{center}
\includegraphics[width=0.24\linewidth]{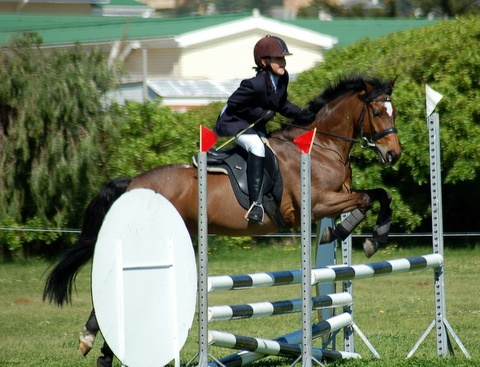} \includegraphics[width=0.24\linewidth]{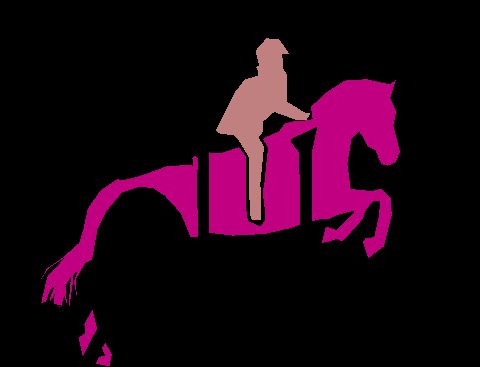} \includegraphics[width=0.24\linewidth]{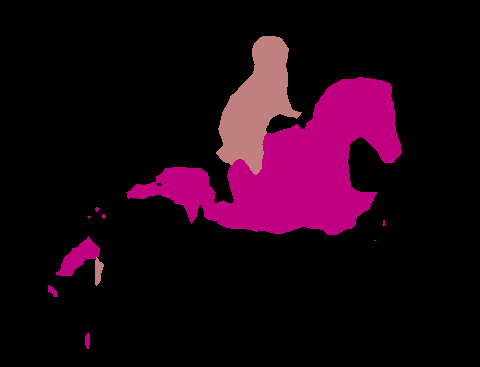} \includegraphics[width=0.24\linewidth]{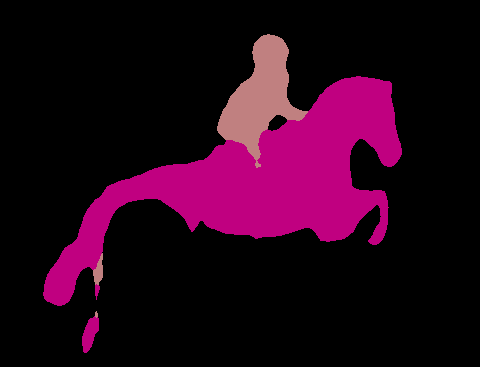}\\
\includegraphics[width=0.24\linewidth]{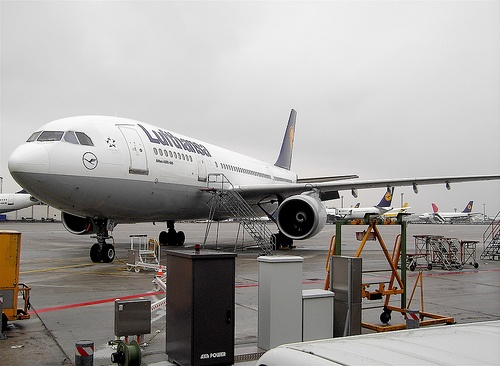} \includegraphics[width=0.24\linewidth]{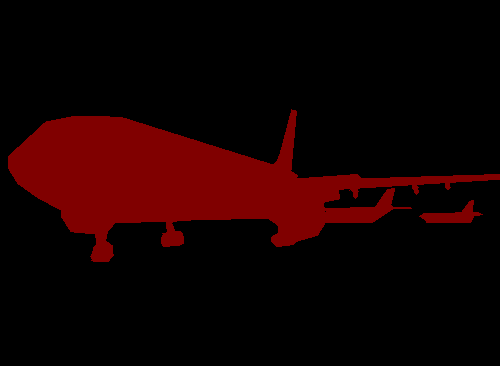} \includegraphics[width=0.24\linewidth]{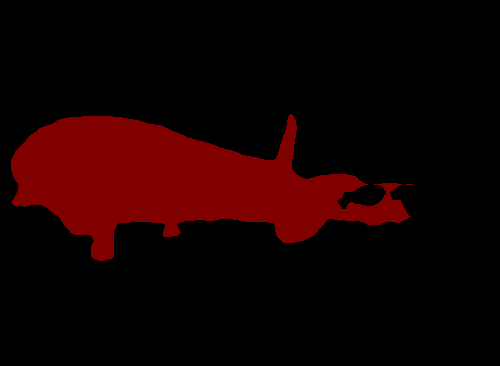} \includegraphics[width=0.24\linewidth]{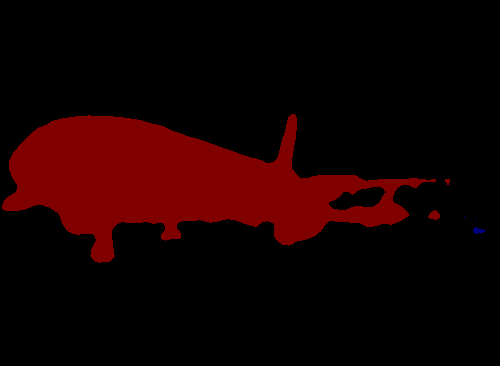}\\
\includegraphics[width=0.24\linewidth]{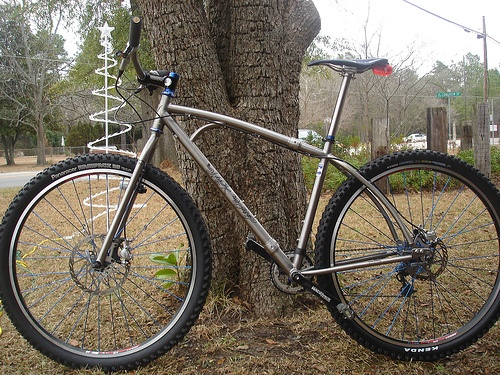} \includegraphics[width=0.24\linewidth]{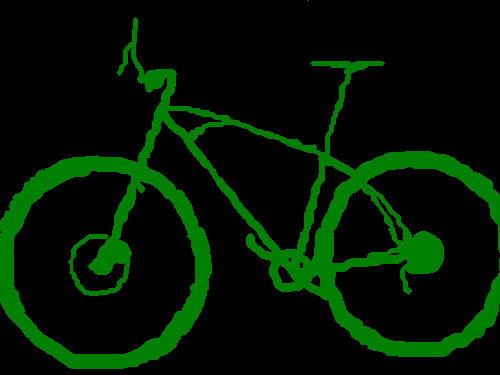} \includegraphics[width=0.24\linewidth]{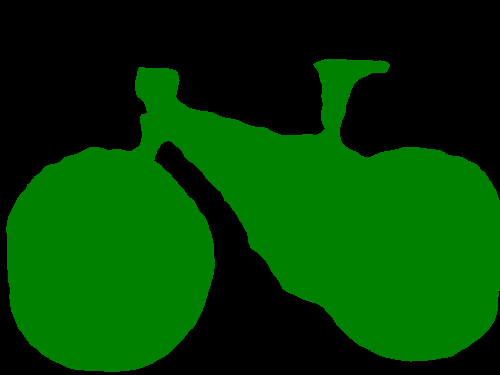} \includegraphics[width=0.24\linewidth]{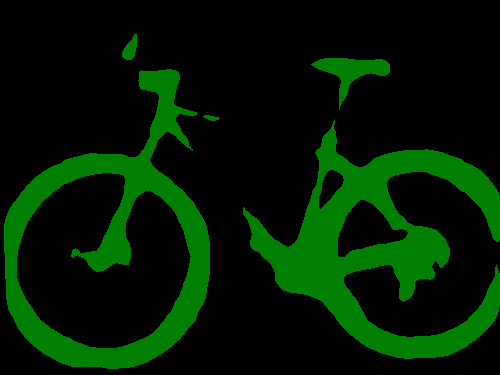}\\
\includegraphics[width=0.24\linewidth]{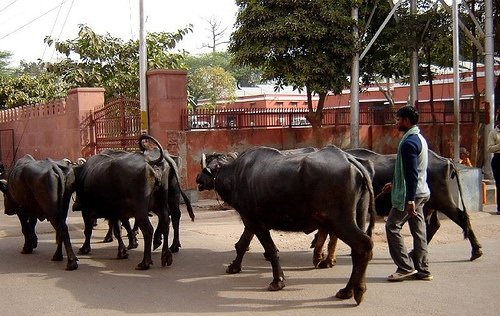} \includegraphics[width=0.24\linewidth]{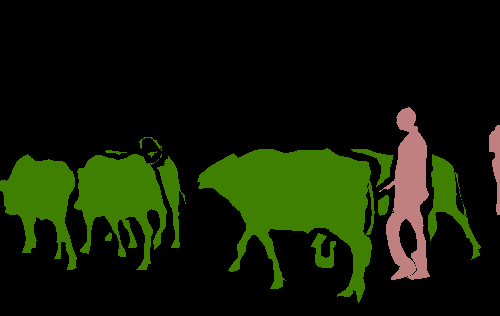} \includegraphics[width=0.24\linewidth]{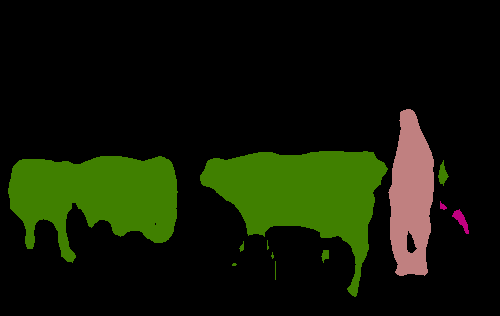} \includegraphics[width=0.24\linewidth]{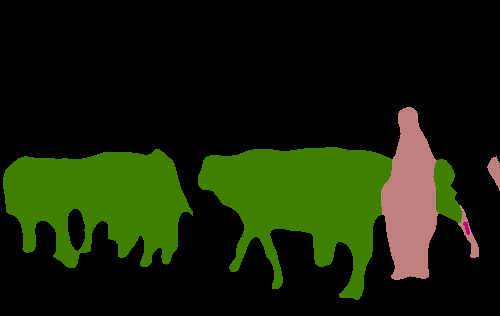}\\
\includegraphics[width=0.24\linewidth]{figures/comparisions/best_results_seg/24_rgb.png} \includegraphics[width=0.24\linewidth]{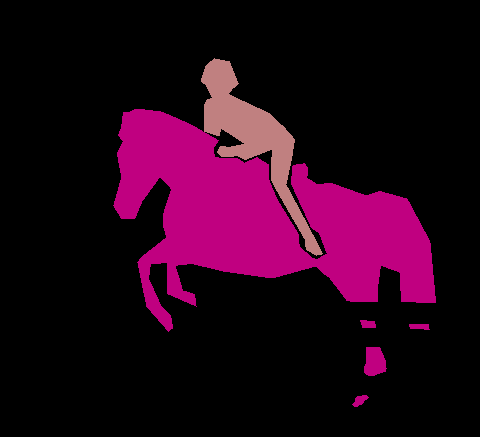} \includegraphics[width=0.24\linewidth]{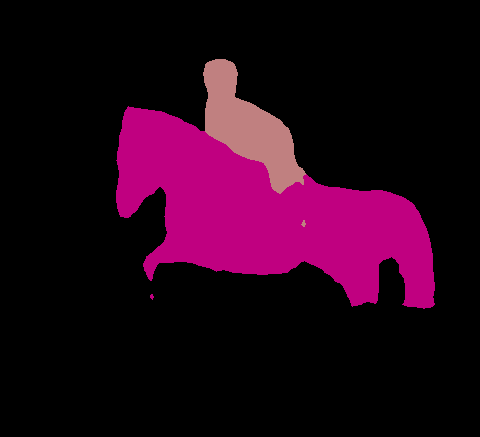} \includegraphics[width=0.24\linewidth]{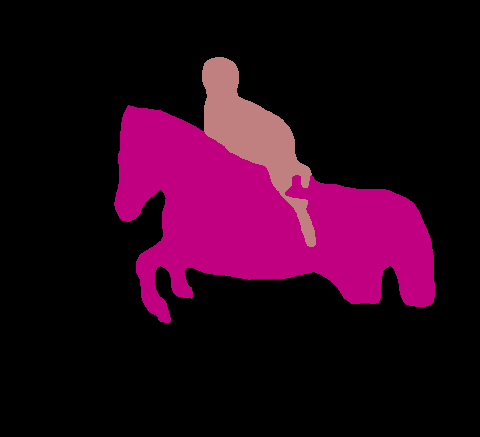}\\
\includegraphics[width=0.24\linewidth]{figures/comparisions/best_results_seg/294_rgb.png} \includegraphics[width=0.24\linewidth]{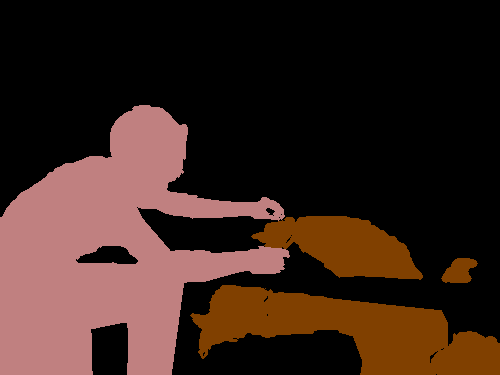} \includegraphics[width=0.24\linewidth]{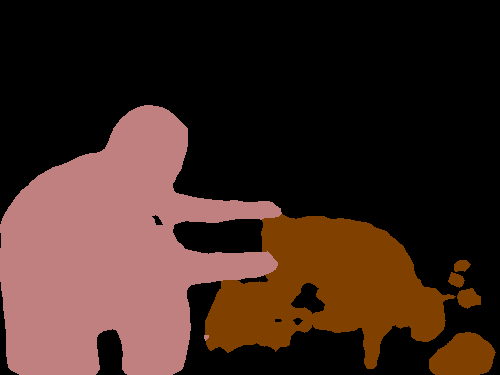} \includegraphics[width=0.24\linewidth]{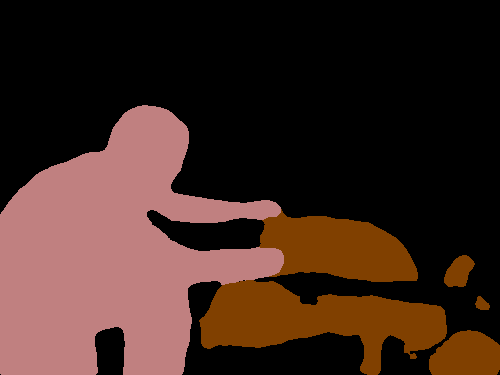}\\
\includegraphics[width=0.24\linewidth]{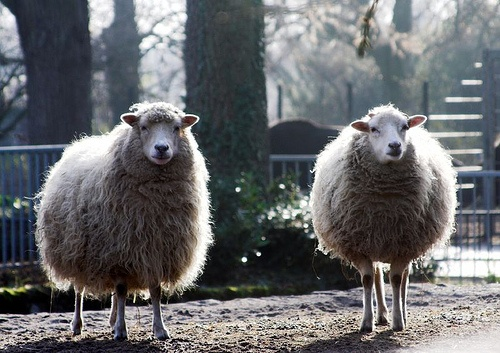} \includegraphics[width=0.24\linewidth]{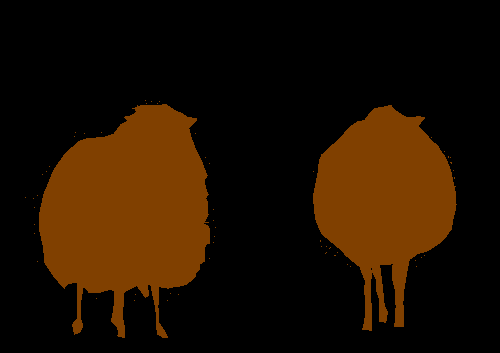} \includegraphics[width=0.24\linewidth]{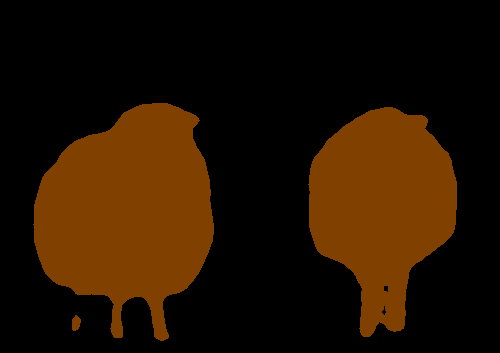} \includegraphics[width=0.24\linewidth]{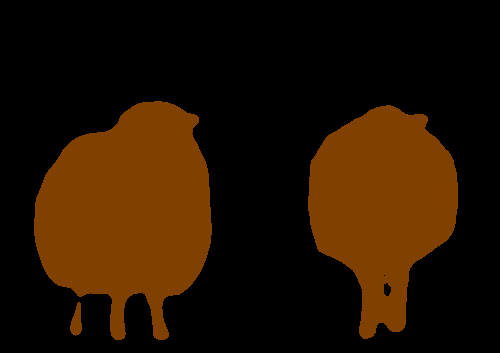}\\
\includegraphics[width=0.24\linewidth]{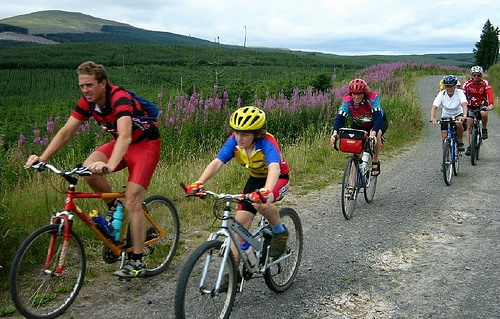} \includegraphics[width=0.24\linewidth]{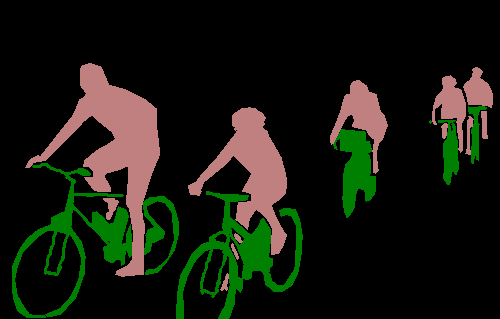} \includegraphics[width=0.24\linewidth]{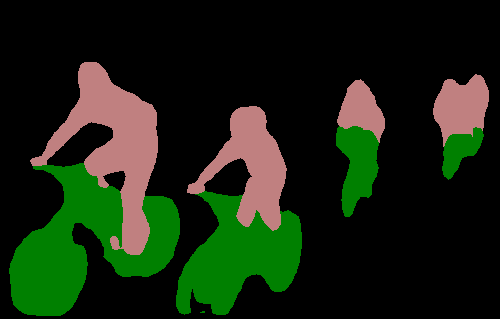} \includegraphics[width=0.24\linewidth]{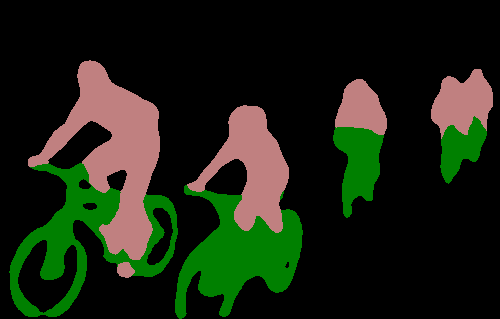}\\
\includegraphics[width=0.24\linewidth]{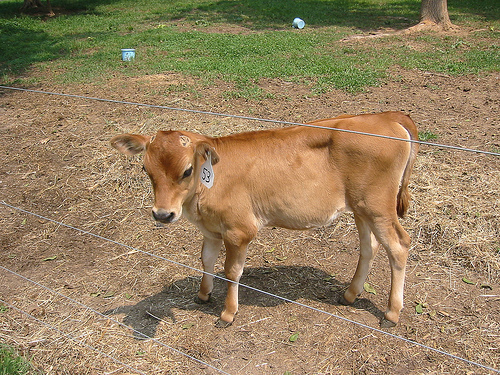} \includegraphics[width=0.24\linewidth]{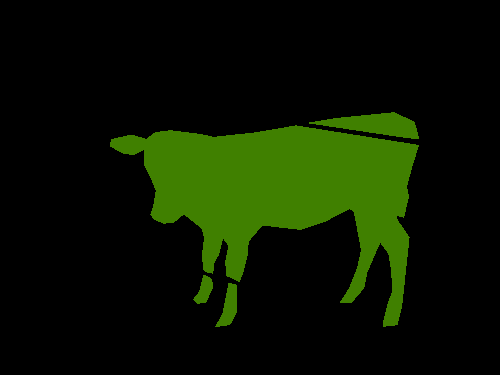} \includegraphics[width=0.24\linewidth]{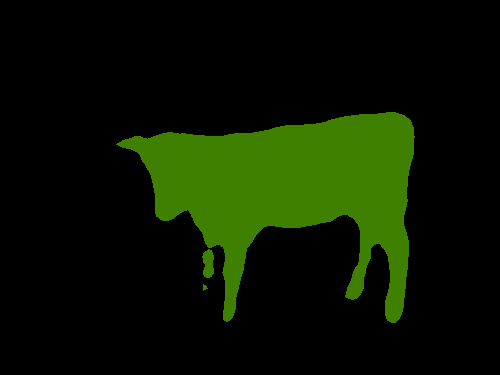} \includegraphics[width=0.24\linewidth]{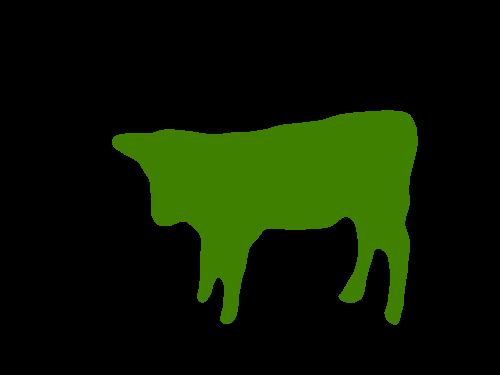}\\
\includegraphics[width=0.24\linewidth]{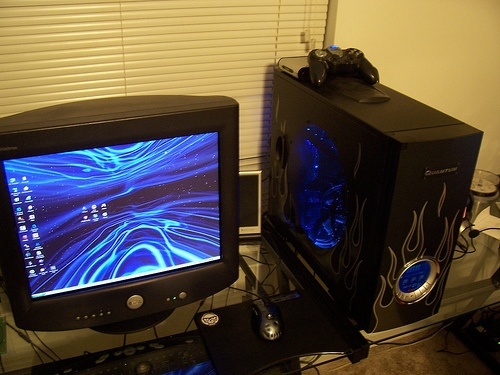} \includegraphics[width=0.24\linewidth]{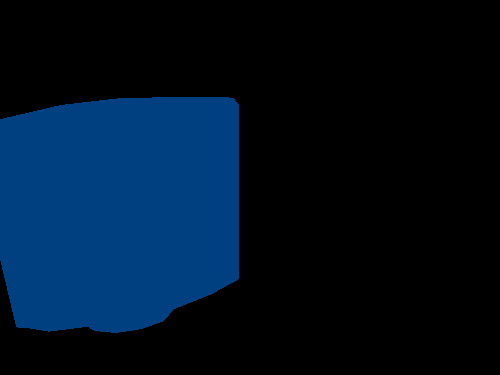} \includegraphics[width=0.24\linewidth]{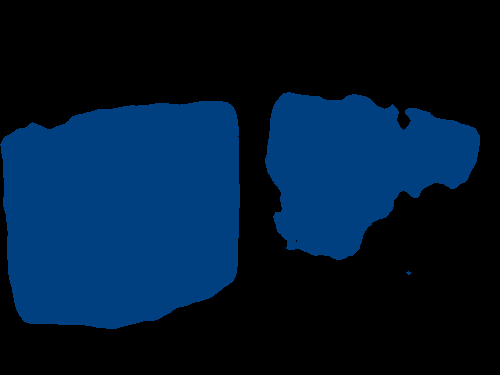} \includegraphics[width=0.24\linewidth]{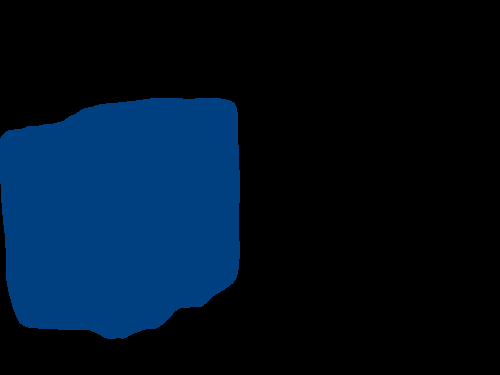}\\
\includegraphics[width=0.24\linewidth]{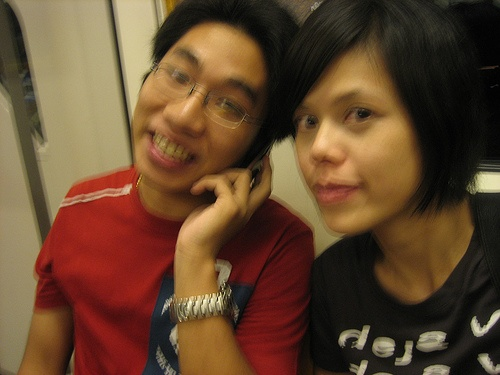} \includegraphics[width=0.24\linewidth]{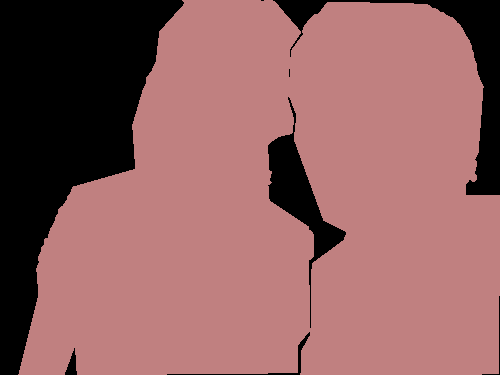} \includegraphics[width=0.24\linewidth]{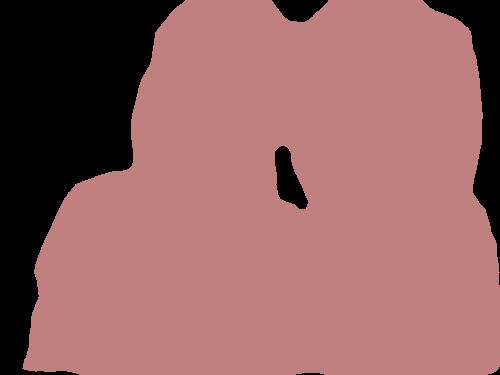} \includegraphics[width=0.24\linewidth]{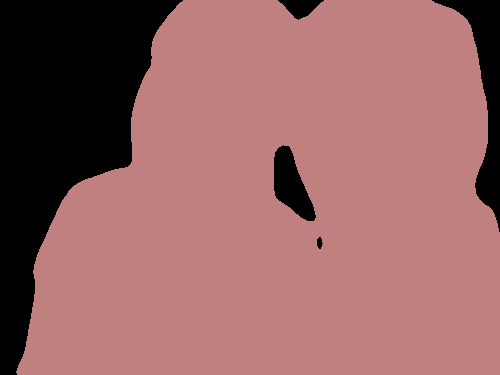}\\
\includegraphics[width=0.24\linewidth]{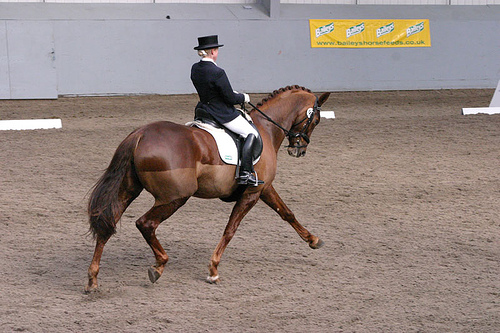} \includegraphics[width=0.24\linewidth]{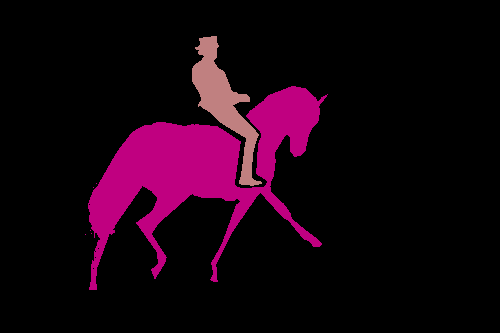} \includegraphics[width=0.24\linewidth]{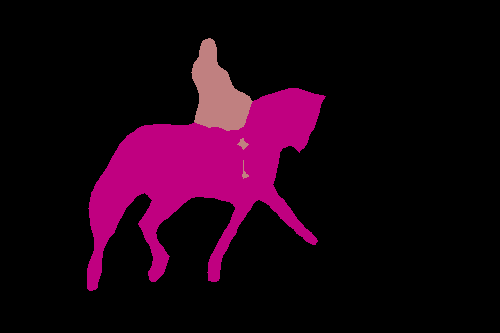} \includegraphics[width=0.24\linewidth]{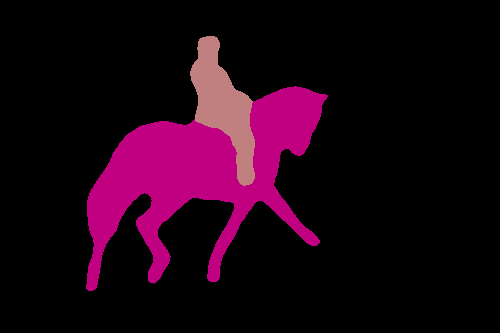}\\
\end{center}

\vspace{-3mm}
\small{(a) Image \hspace{4mm} (b) Ground Truth \hspace{1mm} (c) Unary \hspace{3mm} (d) \texttt{dense-G-CRF}}

\vspace{1mm}
\caption{Qualitative Results of the Semantic Segmentation Task on the \texttt{PASCAL VOC 2012} image segmentation validation dataset. The first column (a) shows the input image, the second column (b) shows the segmentation ground truth,
the third column (c) shows the unary or base network, and the fourth column (d) shows the \texttt{dense-G-CRF} output. It can be seen that the pairwise terms from the pixel embeddings enforce pairwise
consistencies by gathering evidence from other image regions, thus demonstrating better performance. Further, the \texttt{dense-G-CRF} network better captures finer details, and better connects parts of the same object separated by ambiguous image regions.}
\label{fig:visseg}
\end{figure}

We report the results in table \ref{table:seg}.
We report an improvement of $0.3\%$ over the sparse deep G-CRF approach. Using dense-CRF post processing boosts the performance of all methods
consistently. Qualitative improvements are shown in Fig.~\ref{fig:visseg}.

\begin{table}[!h]	
\centering
\begin{tabular}{l|c}
\hline
Method	& mean IoU \\ \hline
Deeplab Cross-Joint \cite{deeplab2} & 73.9 \\ \hline
CRFRNN \cite{crfrnn} & 74.7 \\ \hline
Deeplab Multi-Scale + CRF \cite{edg} & 74.8 \\\hline
Adelaide Context~\cite{Adelaide} & 77.8 \\ \hline
Deep Parsing Network~\cite{DPN} & 77.4 \\ \hline
Deeplab V2 + CRF \cite{deeplabv2} & 79.7 \\\hline
Deeplab-V2 G-CRF Potts \cite{gcrf} & 79.5 \\\hline
Deeplab-V2 G-CRF + CRF Potts \cite{gcrf} & 80.2 \\\hline
\texttt{dense-G-CRF} (Ours)   & 79.8 \\ \hline
\texttt{dense-G-CRF} + CRF (Ours)   & 80.4 \\ \hline
\end{tabular}
\vspace{2mm}
\caption{Semantic segmentation  - mean Intersection Over Union (IOU) accuracy on PASCAL VOC 2012 test.\label{table:seg}}
\end{table}

\subsection{Human Parts Segmentation}
\label{sec:parts}
The human parts segmentation task involves semantic segmentation of human parts. This is another challenging
dense labeling task because of the huge variation in the human pose in different scenarios.
 
\textbf{Dataset.} We use the PASCAL Person Parts dataset introduced in  \cite{parts}. 
This dataset is a subset of the PASCAL VOC 2010 dataset,
with human part segmentations annotated by Chen \etal \cite{parts}, and has humans
in a large variety of poses and scales. 
The dataset contains detailed part segmentations for every person.
As in \cite{graphlstm}, we merge the annotations to obtain six person part classes,
namely the head, torso, upper arms, lower arms, upper legs, and lower legs.
Additionally we have a seventh background class. This dataset has $1716$ training images
and $1817$ testing images. The evaluation
criterion is the pixel intersection-over-union (IOU) metric, averaged across the $7$ classes.

 \textbf{Baselines.} As in the case of semantic segmentation, the state of the art approaches on
 human part segmentation also use fully convolutional networks, sometimes additionally exploiting 
 Long Short Term Memory Units~\cite{graphlstm,parts_lstm}.
For this task, we compare our approach to the following methods:
(a) the deeplab attention to scale network~\cite{ChenYWXY15} which proposes to weigh the features at multiple image scales
with deep supervision, (b) the Auto Zoom network~\cite{HAZN} which adaptively resizes predicted object
instances and parts to the desired scale for refinement of parsed objects, (c) the Local Global LSTM network~\cite{parts_lstm} which 
combines local and global cues via LSTM units, (d) the Graph LSTM network~\cite{graphlstm} which proposes a sophisticated
super-pixel based graph construction approach to generate sequences that are then memorized by LSTM units, (e) the base network
with and without dense CRF post-processing, and (f) the sparse G-CRF Potts model.

\begin{figure}
\begin{center}
\includegraphics[width=0.24\linewidth]{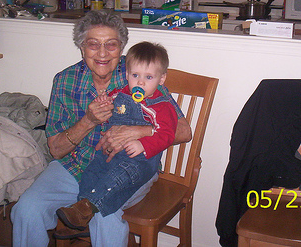} \includegraphics[width=0.24\linewidth]{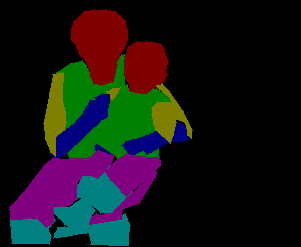} \includegraphics[width=0.24\linewidth]{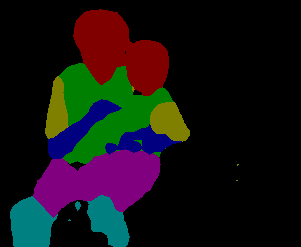} \includegraphics[width=0.24\linewidth]{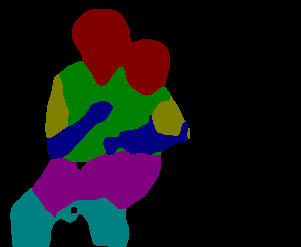}\\
\includegraphics[width=0.24\linewidth]{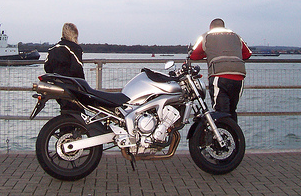} \includegraphics[width=0.24\linewidth]{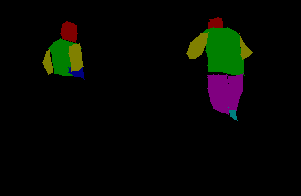} \includegraphics[width=0.24\linewidth]{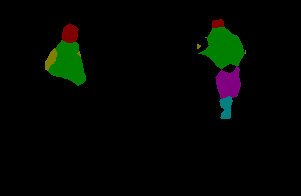} \includegraphics[width=0.24\linewidth]{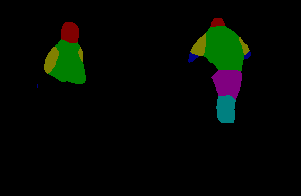}\\
\includegraphics[width=0.24\linewidth]{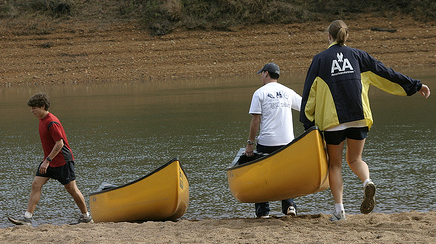} \includegraphics[width=0.24\linewidth]{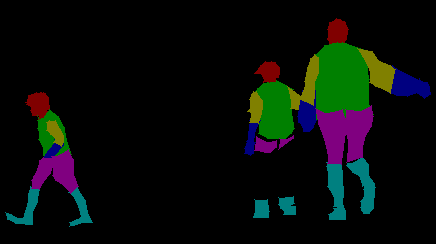} \includegraphics[width=0.24\linewidth]{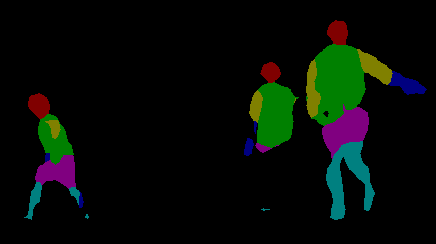} \includegraphics[width=0.24\linewidth]{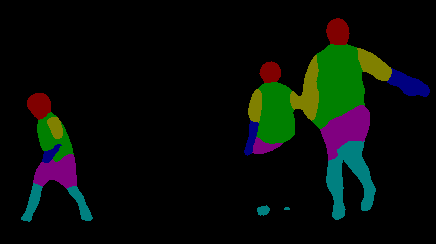}\\
\includegraphics[width=0.24\linewidth]{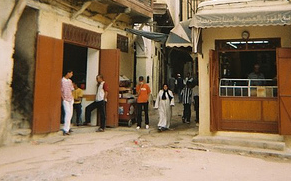} \includegraphics[width=0.24\linewidth]{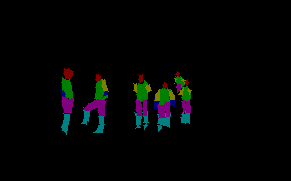} \includegraphics[width=0.24\linewidth]{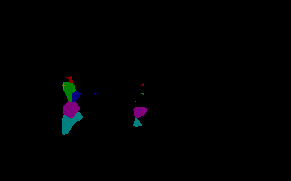} \includegraphics[width=0.24\linewidth]{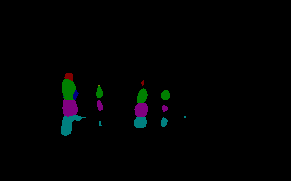}\\
\includegraphics[width=0.24\linewidth]{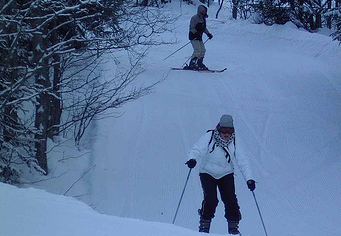} \includegraphics[width=0.24\linewidth]{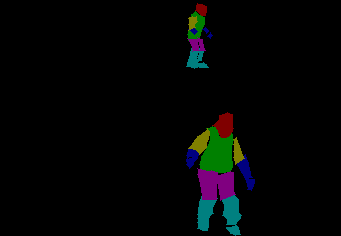} \includegraphics[width=0.24\linewidth]{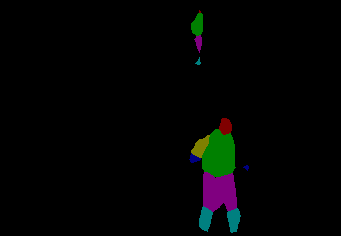} \includegraphics[width=0.24\linewidth]{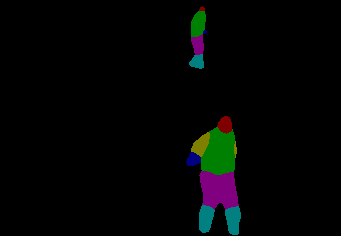}\\
\includegraphics[width=0.24\linewidth]{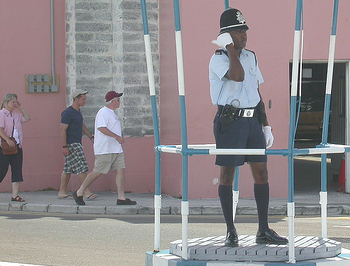} \includegraphics[width=0.24\linewidth]{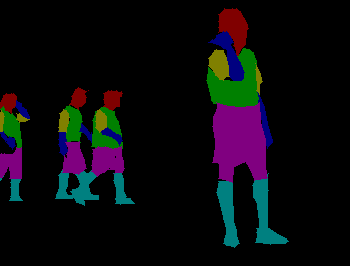} \includegraphics[width=0.24\linewidth]{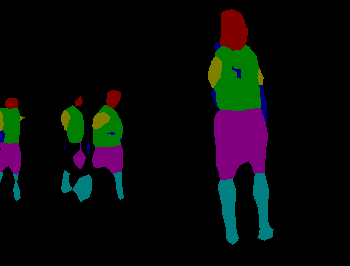} \includegraphics[width=0.24\linewidth]{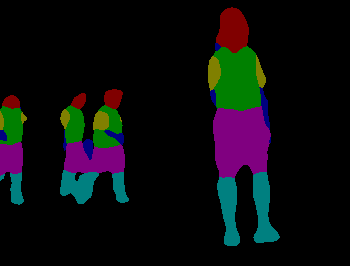}\\
\includegraphics[width=0.24\linewidth]{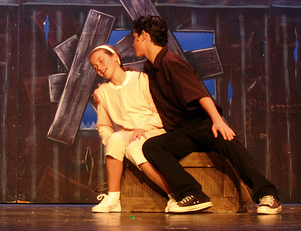} \includegraphics[width=0.24\linewidth]{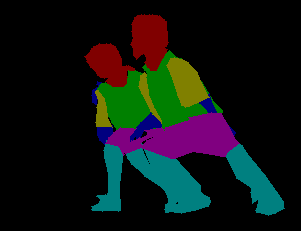} \includegraphics[width=0.24\linewidth]{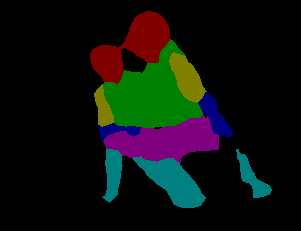} \includegraphics[width=0.24\linewidth]{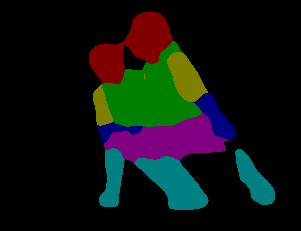}\\
\includegraphics[width=0.24\linewidth]{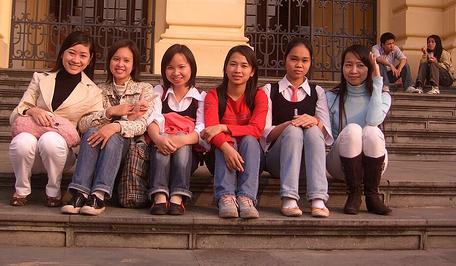} \includegraphics[width=0.24\linewidth]{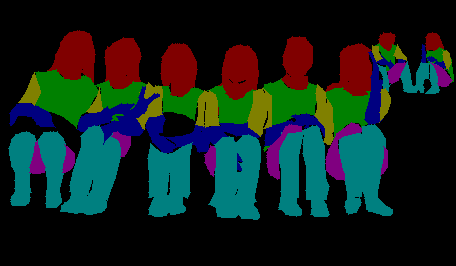} \includegraphics[width=0.24\linewidth]{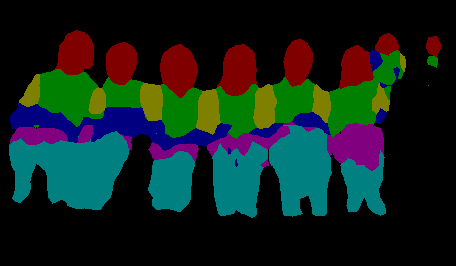} \includegraphics[width=0.24\linewidth]{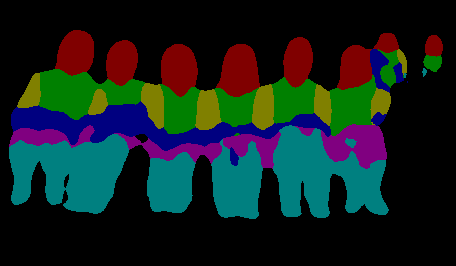}\\
\includegraphics[width=0.24\linewidth]{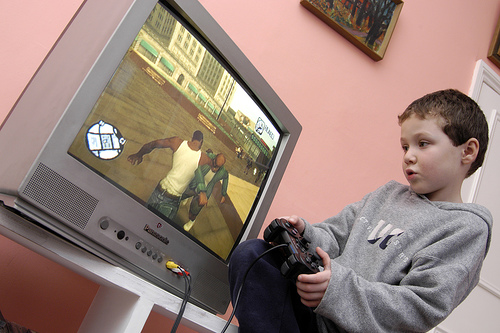} \includegraphics[width=0.24\linewidth]{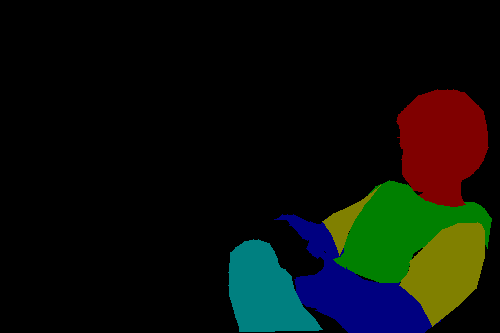} \includegraphics[width=0.24\linewidth]{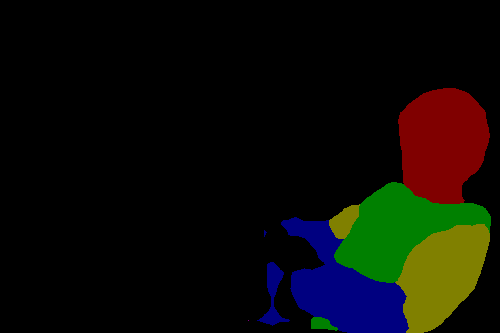} \includegraphics[width=0.24\linewidth]{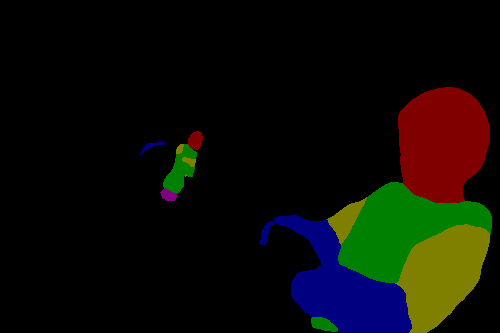}\\
\includegraphics[width=0.24\linewidth]{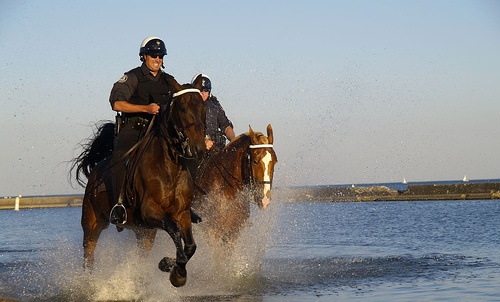} \includegraphics[width=0.24\linewidth]{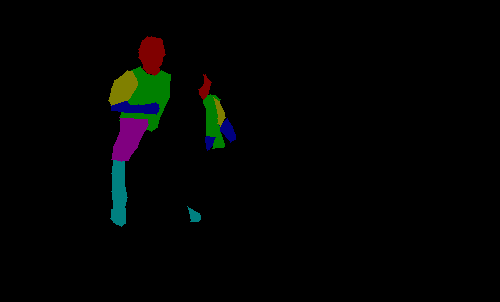} \includegraphics[width=0.24\linewidth]{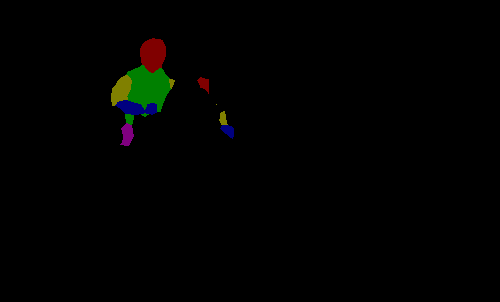} \includegraphics[width=0.24\linewidth]{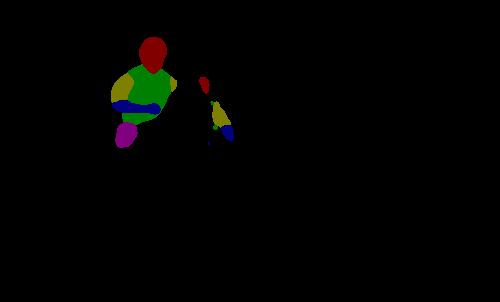}\\
\includegraphics[width=0.24\linewidth]{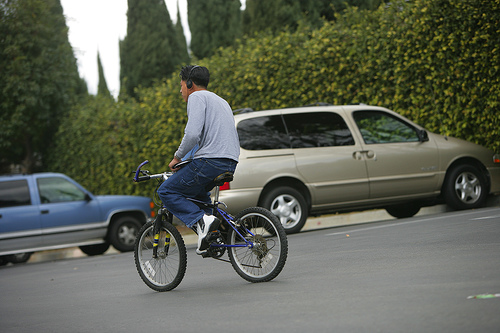} \includegraphics[width=0.24\linewidth]{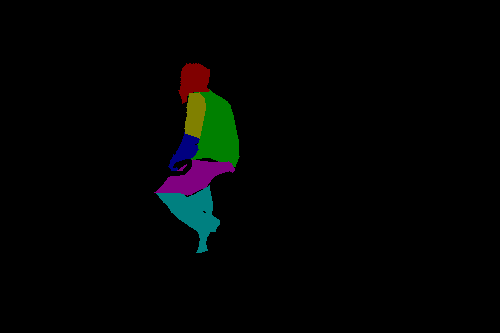} \includegraphics[width=0.24\linewidth]{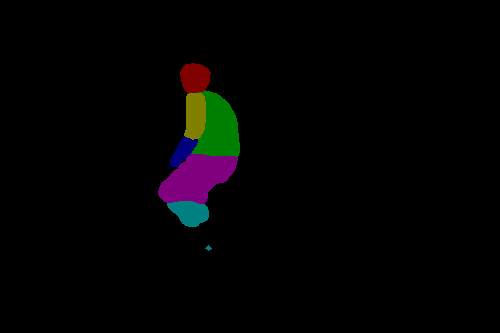} \includegraphics[width=0.24\linewidth]{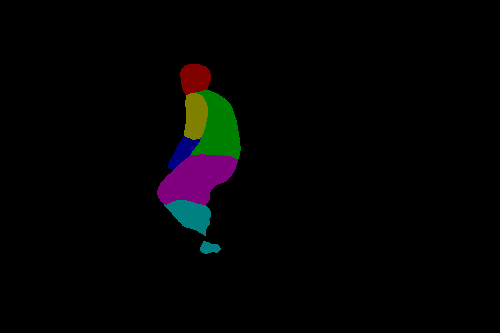}\\
\includegraphics[width=0.24\linewidth]{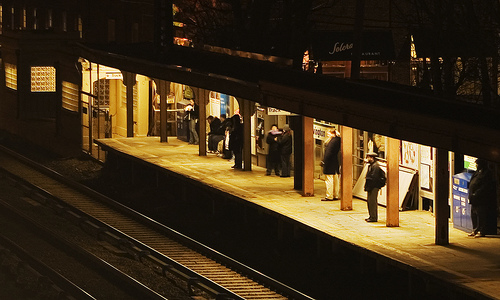} \includegraphics[width=0.24\linewidth]{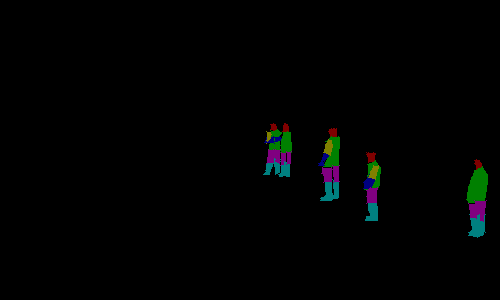} \includegraphics[width=0.24\linewidth]{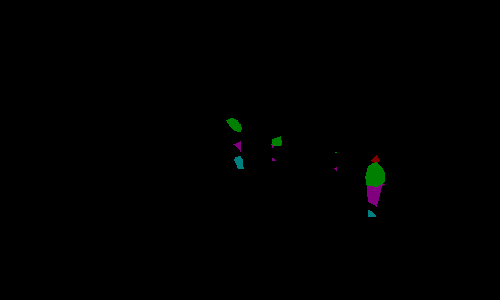} \includegraphics[width=0.24\linewidth]{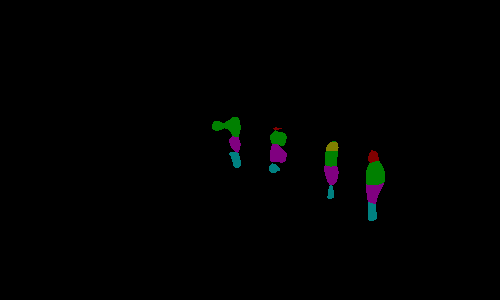}\\
\end{center}

\vspace{-3mm}
\small{(a) Image \hspace{4mm} (b) Ground Truth \hspace{1mm} (c) Unary \hspace{3mm} (d) \texttt{dense-G-CRF}}

\caption{Qualitative Results of the Part Segmentation Task on the PASCAL Parts validation dataset. The first column (a) shows the input image, the second column (b) shows the part ground truth,
the third column (c) shows the unary or base network, and the fourth column (d) shows the \texttt{dense-G-CRF} output. Our method captures the boundaries better than the base network.}
\label{fig:visparts}
\end{figure}

We report the results in table \ref{table:parts}. For our \texttt{dense-G-CRF} method, we use an embedding dimension of $128$. 
Please note that while the previous state of the art approach deeplab-v2 achieves $64.94$ with dense-CRF post-processing,
we outperform it by $1.16\%$ mean IoU without using dense-CRF post-processing. Additionally, we outperform the Deeplab-V2 G-CRF Potts
baseline from \cite{gcrf} by $0.9\%$. We show qualitative results in Fig.~\ref{fig:visparts}.
\begin{table}[!h]
\centering
\begin{tabular}{l|c}
\hline
Attention ~\cite{ChenYWXY15} & 56.39 \\\hline
Auto Zoom~\cite{HAZN}  & 57.54 \\\hline
LG-LSTM ~\cite{parts_lstm}  &57.97 \\\hline
Graph LSTM ~\cite{graphlstm} & 60.16 \\\hline
Deeplab-V2 ~\cite{deeplabv2} & 64.40 \\\hline
Deeplab-V2-CRF~\cite{deeplabv2} & 64.94 \\\hline
Deeplab-V2 G-CRF Potts \cite{gcrf} & 65.21 \\\hline
\texttt{dense-G-CRF} (Ours) ~\cite{deeplabv2} & 66.10 \\\hline
\end{tabular}
\vspace{2mm}
\caption{Part segmentation - mean Intersection-Over-Union accuracy on the PASCAL Parts dataset of \cite{parts}. \label{table:parts}} \end{table}
 
\subsection{Saliency Estimation}
\label{sec:sal}
The saliency estimation problem intends to discover the most interesting regions in an image. It is also
posed as a dense labeling task where the objective is to assign a score to each pixel which is proportional
to the magnitude of attention that an observer would give to the pixel. This task is very hard to evaluate
since the magnitude of attention is a subjective quantity. Standard benchmarks work around this difficulty
by having the dataset annotated by a set of annotators, and then averaging their annotations to estimate an
average consensus. These scores are normalized and thresholded
to give a binary classification to each pixel indicating whether the image region is interesting or not.

\textbf{Dataset.} We use a training protocol similar to \cite{ubernet} for this task.
More specifically, we use the MSRA-10K saliency dataset introduced in \cite{WangLLLS15} for training,
and evaluate our performance on the PASCAL-S dataset introduced in \cite{LiHKRY14}, as well as
the HKU-IS dataset from \cite{LiY15}. The
MSRA-10K dataset contains $10000$ images with annotated pixel-wise
segmentation masks for salient objects. The Pascal-S saliency dataset contains pixel-wise saliency
estimates in the $\left[0,1\right]$ range for $850$ images. As suggested in  
\cite{LiHKRY14}, we threshold the saliency values at $0.5$ to obtain the binary masks. The
HKU-IS dataset has $4447$ images, and is known for low-contrast and multiple salient objects in each image.
The evaluation criterion is the 
maximal F-Measure as in ~\cite{ubernet,LiHKRY14}.

\textbf{Baselines.} 
Our baselines for the saliency estimation task include (a) BSCA ~\cite{QinLXW15}
which is a classical approach based on a dynamic evolution model called the cellular automata and Bayesian inference,
(b) the Local Estimation and Global Search (LEGS) framework~\cite{WangLRY15} which is a deep learning based approach
that combines local and global cues,
(c) the multi-context network~\cite{ZhaoOLW15} that captures both global and local context in the input image,
(d) the multiscale deep features network~\cite{LiY15} which proposes a multi-resolution network architecture,
a refinement strategy and aggregates saliency maps at different levels, (e) the deep contrast learning networks~\cite{saliencycvpr16},
which proposes a network structure that better exploits object boundaries to improve saliency estimation and additionally uses a 
fully connected CRF model, (f)
the Ubernet architecture~\cite{ubernet} which demonstrates that sharing parameters for mutually symbiotic tasks can help
improve overall performance of these tasks, (g) our base network, i.e. deeplab-v2, and (h) the sparse G-CRF Potts model alongside the base network.

\begin{figure}
\begin{center}
\includegraphics[width=0.24\linewidth]{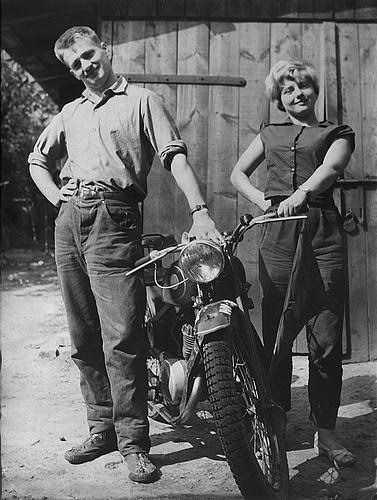} \includegraphics[width=0.24\linewidth]{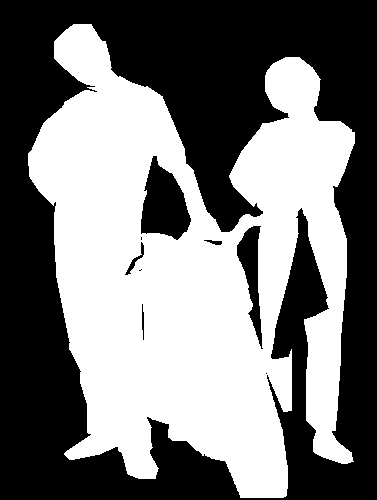} \includegraphics[width=0.24\linewidth]{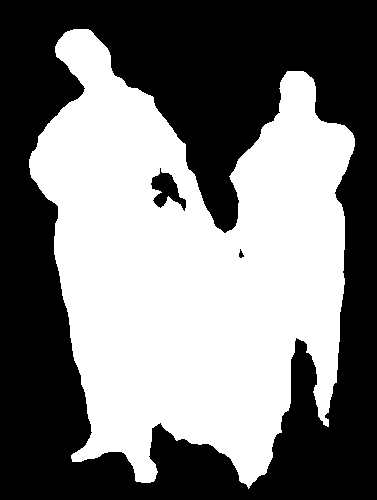} \includegraphics[width=0.24\linewidth]{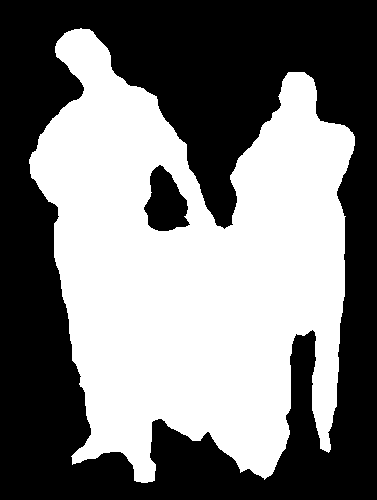}\\
\includegraphics[width=0.24\linewidth]{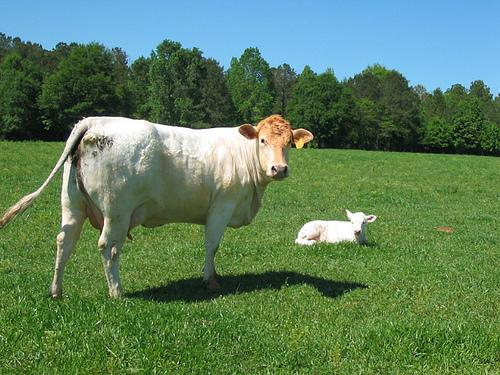} \includegraphics[width=0.24\linewidth]{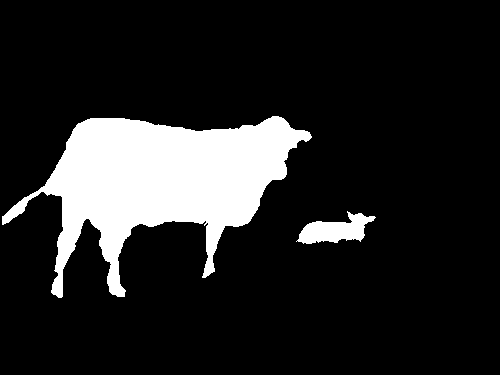} \includegraphics[width=0.24\linewidth]{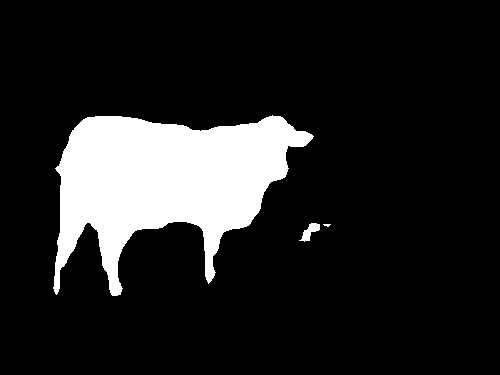} \includegraphics[width=0.24\linewidth]{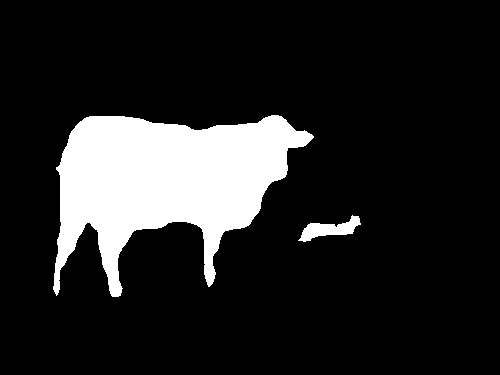}\\
\includegraphics[width=0.24\linewidth]{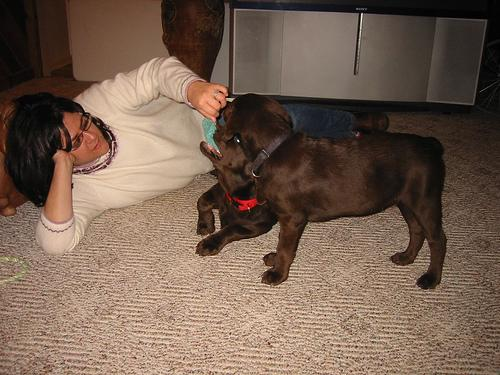} \includegraphics[width=0.24\linewidth]{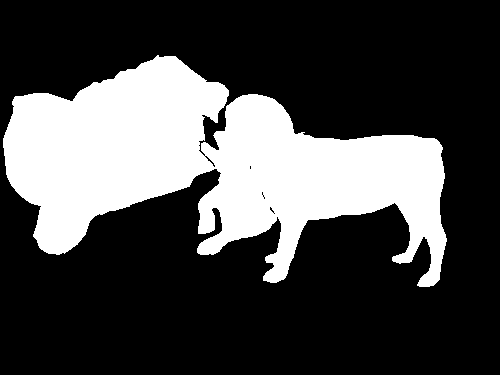} \includegraphics[width=0.24\linewidth]{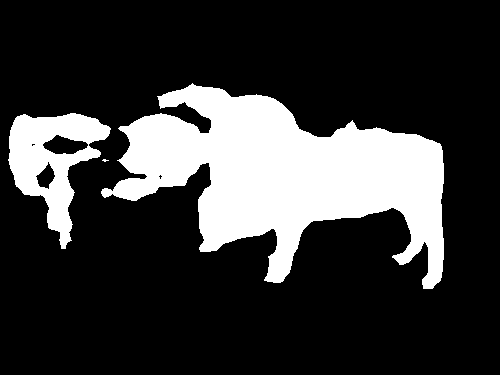} \includegraphics[width=0.24\linewidth]{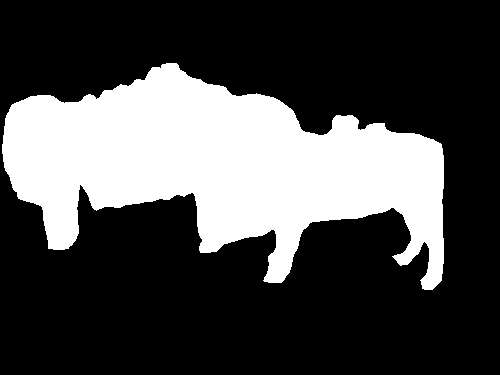}\\
\includegraphics[width=0.24\linewidth]{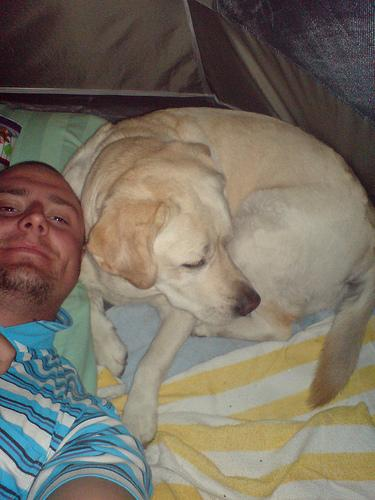} \includegraphics[width=0.24\linewidth]{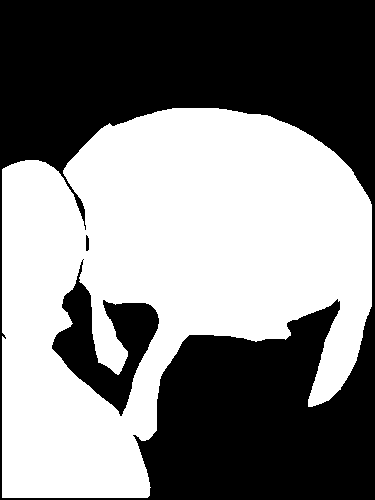} \includegraphics[width=0.24\linewidth]{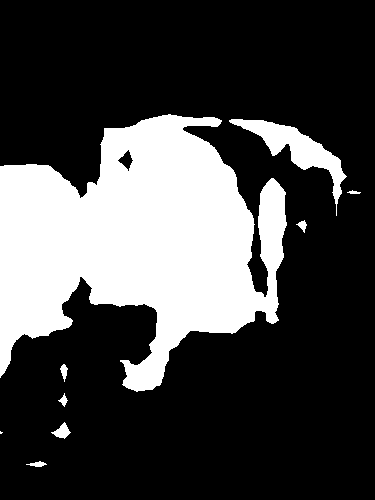} \includegraphics[width=0.24\linewidth]{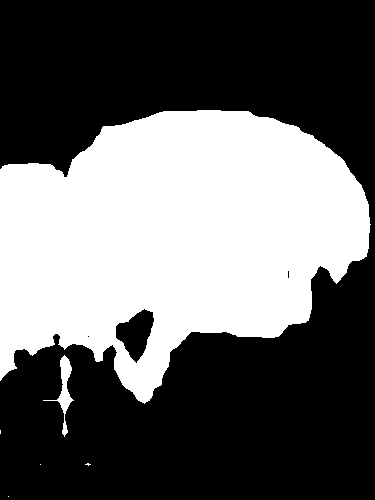}\\
\includegraphics[width=0.24\linewidth]{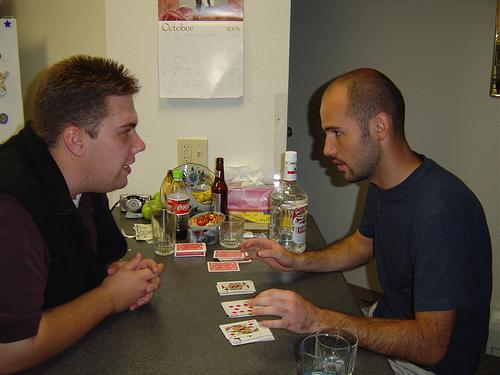} \includegraphics[width=0.24\linewidth]{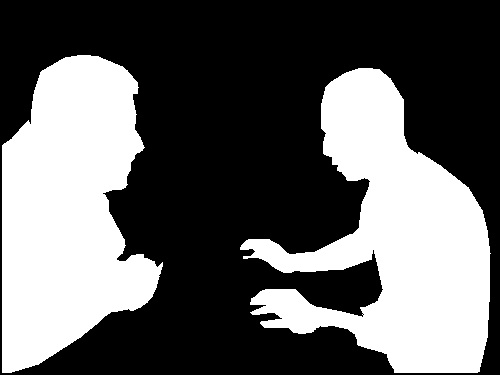} \includegraphics[width=0.24\linewidth]{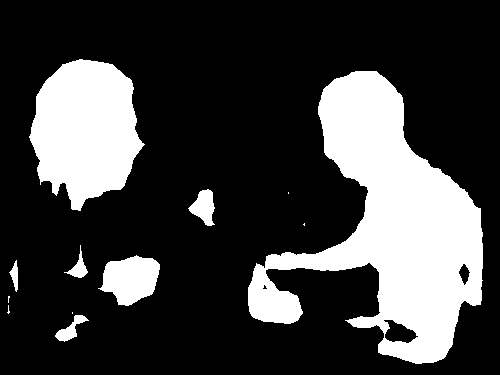} \includegraphics[width=0.24\linewidth]{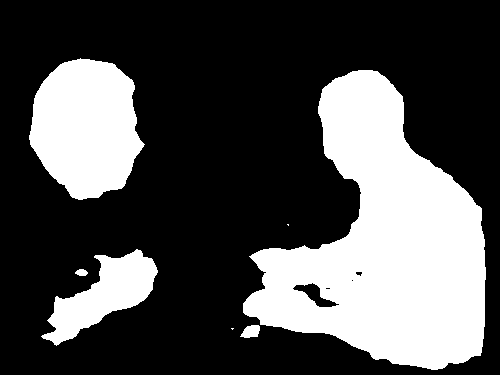}\\
\includegraphics[width=0.24\linewidth]{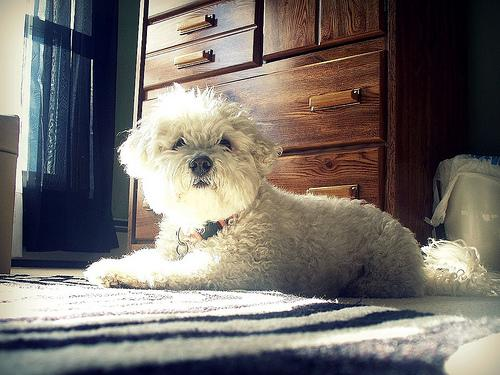} \includegraphics[width=0.24\linewidth]{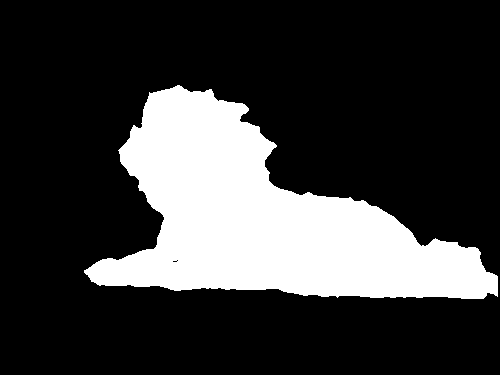} \includegraphics[width=0.24\linewidth]{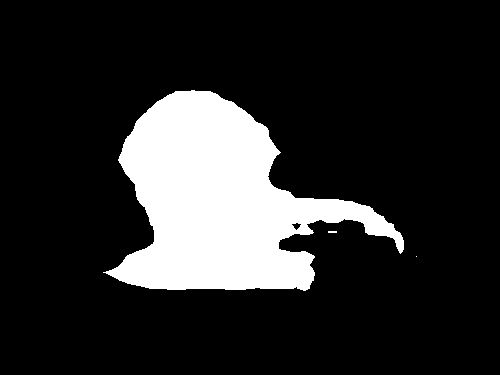} \includegraphics[width=0.24\linewidth]{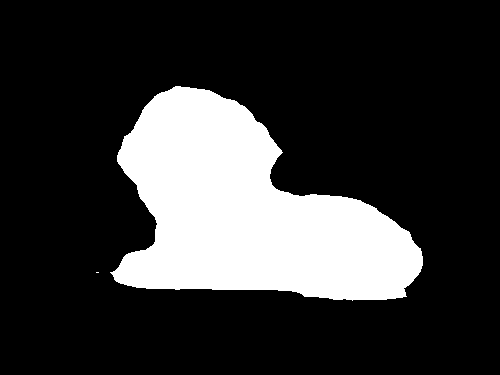}\\
\includegraphics[width=0.24\linewidth]{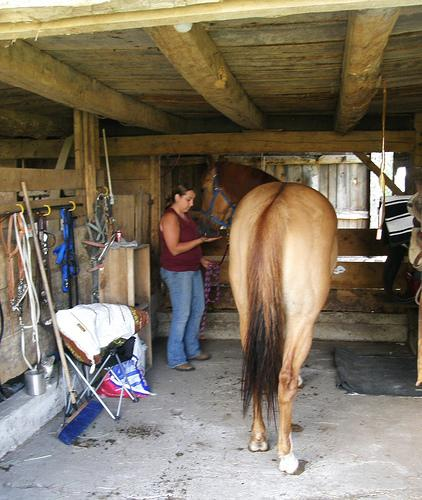} \includegraphics[width=0.24\linewidth]{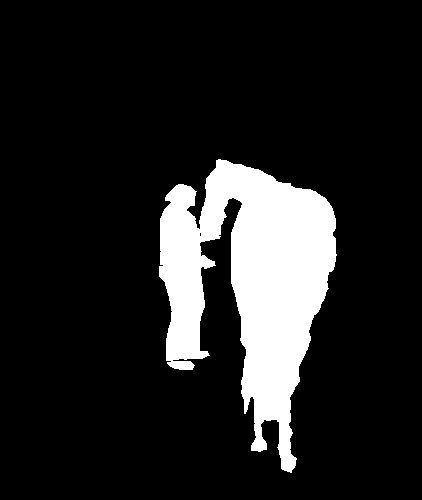} \includegraphics[width=0.24\linewidth]{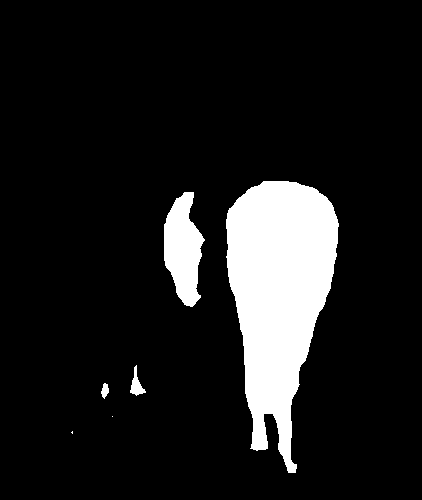} \includegraphics[width=0.24\linewidth]{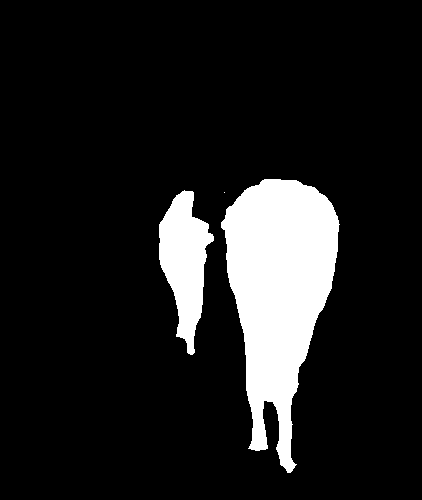}\\
\end{center}

\vspace{-3mm}
\small{(a) Image \hspace{4mm} (b) Ground Truth \hspace{1mm} (c) Unary \hspace{3mm} (d) \texttt{dense-G-CRF}}

\vspace{1mm}
\caption{Qualitative Results of the Visual Saliency Estimation Task on the PASCAL-S dataset. The first column (a) shows the input image, the second column (b) shows the saliency ground truth,
the third column (c) shows the unary or base network, and the fourth column (d) shows the \texttt{dense-G-CRF} output. It can be seen that the pairwise terms help improve the base network performance
significantly.}
\label{fig:vissal}
\end{figure}

The results are tabulated in table \ref{table:saliency}. Our approach yields a significant $0.3$ point improvement in
maximal F-score over the other previous state of the art Ubernet approach.
For our \texttt{dense-G-CRF} method, we use an embedding dimension of $16$. We show some qualitative results in Fig.~\ref{fig:vissal}.

\begin{table}[!h]
\centering
\begin{tabular}{l|r|r}		
\hline	
Method	&  PASCAL-S & HKU-IS\\\hline
BSCA \cite{QinLXW15} & 0.666 & 0.723\\\hline
LEGS \cite{WangLRY15} & 0.752 & 0.770 \\\hline
MC  \cite{ZhaoOLW15} & 0.740 & 0.798 \\\hline
MDF  \cite{LiY15} & 0.764 & 0.861 \\\hline
FCN  \cite{saliencycvpr16} & 0.793 & 0.867 \\\hline
DCL   \cite{saliencycvpr16} & 0.815 & 0.892\\\hline
DCL + CRF \cite{saliencycvpr16} & 0.822 & 0.904\\\hline
Ubernet 1-Task \cite{ubernet} & 0.835 & - \\\hline
Deeplab-v2 \cite{deeplabv2} & 0.859 & 0.916 \\\hline
Deeplab-V2 G-CRF Potts \cite{gcrf} & 0.861 & 0.914 \\\hline
\texttt{dense-G-CRF} (Ours) & 0.864 & 0.919 \\\hline
\end{tabular}
\vspace{2mm}
\caption{Saliency estimation results: we report the Maximal F-measure (MF) on the PASCAL Saliency dataset of \cite{LiHKRY14}, and the HKU-IS dataset of
\cite{LiY15}.\label{table:saliency}
}
\end{table}

\section{Conclusions and Future Work}
In this work we propose a fully-connected G-CRF model for end-to-end training of
deep architectures. Our model expresses the pairwise interactions in the G-CRF model
as a low-rank matrix composed via dot-products of low-dimensional pixel embeddings. This allows us to perform exact inference in fully connected models, while using efficient and low-memory conjugate gradient solvers that exploit the low-rank nature of the resulting system.  
Our implementation is fully GPU based, and implemented using the \emph{Caffe} library.
Our experimental evaluation indicates consistent improvements over the state of the
art approaches on three challenging public benchmarks for semantic segmentation,
human parts segmentation and saliency estimation.
In the future, we intend to exploit
this framework on other dense labeling tasks, as well as regression tasks, such as depth estimation,  image denoising and normal estimation, which can be naturally handled by our model thanks to its continuous nature.

{\small
\bibliographystyle{ieee}
\bibliography{egbib}
}

\end{document}